\documentclass[conference]{IEEEtran}
\IEEEoverridecommandlockouts

\usepackage{cite}
\usepackage{amsmath,amssymb,amsfonts, mathtools, array}

\usepackage{graphicx}
\usepackage{textcomp}
\usepackage{xcolor}
\usepackage{algorithm}
\usepackage{algpseudocode}
\usepackage{caption}
\usepackage{subcaption}
\usepackage{tabularx}
\usepackage{float}
\usepackage{lipsum,mathtools}
\usepackage{multirow}
\usepackage{dirtytalk}
\usepackage{amssymb, nccmath}
\usepackage[numbers, sort&compress]{natbib}
\def\BibTeX{{\rm B\kern-.05em{\sc i\kern-.025em b}\kern-.08em
    T\kern-.1667em\lower.7ex\hbox{E}\kern-.125emX}}
    \IEEEaftertitletext{\vspace{-2\baselineskip}}
    \usepackage[left=0.57in,right=0.57in,top=0.7in,bottom=0.95in]{geometry}
\begin{document}
\title{FMLFS: A federated multi-label feature selection based on information theory in IoT environment}
\author{\IEEEauthorblockN{Afsaneh Mahanipour}
\IEEEauthorblockA{\textit{Department of Computer Science} \\
\textit{University of Kentucky}\\
Lexington, KY, USA \\
ama654@uky.edu}
\and
\IEEEauthorblockN{Hana Khamfroush}
\IEEEauthorblockA{\textit{Department of Computer Science} \\
\textit{University of Kentucky}\\
Lexington, KY, USA \\
khamfroush@cs.uky.edu}
}
\maketitle
\begin{abstract}
In certain emerging applications such as health monitoring wearable and traffic monitoring systems, Internet-of-Things (IoT) devices generate or collect a huge amount of multi-label datasets. Within these datasets, each instance is linked to a set of labels. The presence of noisy, redundant, or irrelevant features in these datasets, along with the curse of dimensionality, poses challenges for multi-label classifiers. Feature selection (FS) proves to be an effective strategy in enhancing classifier performance and addressing these challenges. Yet, there is currently no existing distributed multi-label FS method documented in the literature that is suitable for distributed multi-label datasets within IoT environments. This paper introduces FMLFS, the first federated multi-label feature selection method. Here, mutual information between features and labels serves as the relevancy metric, while the correlation distance between features, derived from mutual information and joint entropy, is utilized as the redundancy measure. Following aggregation of these metrics on the edge server and employing Pareto-based bi-objective and crowding distance strategies, the sorted features are subsequently sent back to the IoT devices. The proposed method is evaluated through two scenarios: 1) transmitting reduced-size datasets to the edge server for centralized classifier usage, and 2) employing federated learning with reduced-size datasets. Evaluation across three metrics - performance, time complexity, and communication cost - demonstrates that FMLFS outperforms five other comparable methods in the literature and provides a good trade-off on three real-world datasets.
\end{abstract}

\vspace{-1mm}
\section{Introduction}
\vspace{-2mm}
With the development of emerging science and technologies such as Internet-of-Things (IoT), smart healthcare, and intelligent transportation, we are entering the era of big data, where a huge amount of data has been generated daily. In many cases, these collected data may contain irrelevant, noisy, or redundant features. The presence of such features in these data not only leads to increased complexity and execution time of learning models but also significantly impacts their performance \cite{mahanipour2023multimodal}.

Data pre-processing methods can be used to tackle these issues effectively. Among these methods, feature selection (FS) techniques select relevant and informative features from original ones without changing them unlike other dimensional reduction methods like principal component analysis. FS procedure reduces data dimensions, lowers computational costs and storage requirements, while also improving learning model performance \cite{zebari2020comprehensive}. In some real-world applications across different domains such as text categorization, image recognition, and gene prediction, each instance is associated with a set of labels. Such datasets are known as multi-label data. Just like single-label data, multi-label data also face the challenge of high dimensionality. Consequently, effective multi-label feature selection approaches become crucial in addressing this dimensionality problem \cite{hu2022feature}.

On the other hand, the processing of these vast amounts of collected data by intelligent devices, particularly within IoT networks, is essential for extracting meaningful insights about the environment. Traditionally, these datasets were transmitted to cloud servers. However, due to the need for real-time responses and concerns about privacy, many IoT applications now restrict the transfer of data to the cloud. Instead, the data needs to be processed either locally or at the edge to address these requirements \cite{nishio2019client}. If all non-pre-processed local datasets are transferred to an edge server, it will result in increased communication costs between end-user devices/clients and the edge server, leading to delays in processing. Moreover, conducting local processing on non-pre-processed local datasets using distributed machine learning models like federated learning (FL) algorithm would increase the complexity and execution time of these models.

Therefore, to select informative features from distributed multi-label datasets on clients, a collaborative multi-label feature selection method is needed. There are several centralized multi-label FS approaches in the literature that are encountering challenges in these environments. For instance, if each client's data is independently fed to a feature selection process, the issue of feature selection bias may lead to inaccurate and non-robust results. Therefore, a distributed FS like federated feature selection (FFS) procedure should be applied to multi-label FS. According to our best knowledge, this is the first work that investigates federated multi-label FS method. In this study, we compute the mutual information between features and class labels, as well as the correlation distance between features, obtained by subtracting mutual information from joint entropy between features in each client. The proposed method evaluates both the relevance between features and class labels, as well as the redundancy among features base on these two metrics. Next, these values are transmitted to the edge server for aggregation, where they serve as two objectives in a Pareto-based bi-objective strategy. The features are then ranked based on the combination of their Pareto front number and their crowding distance. Then, the ranked features are returned to each client. Consequently, by using a smaller number of features and reducing the data size, the performance of the learning algorithm can be enhanced. This leads to accelerated machine learning (ML) models and reduced complexity. Moreover, it aids in minimizing communication costs when transmitting datasets to edge servers. The main novelties of the proposed method can be summarized as follows:

\begin{itemize}
\item Proposing the first federated multi-label feature selection method
\item Employing information theory-based concepts as two objectives within a Pareto-based bi-objective strategy
\item Utilizing federated learning algorithm as a multi-label classifier for the first time
\item Comparing the proposed method with five other centralized multi-label FS methods on three datasets from three application domains
\end{itemize}
\vspace{-2mm}
\section{RELATED WORKS}
Previous studies mostly focused on centralized multi-label feature selection. Limited research has explored federated feature selection for single-label datasets, and none has investigated federated multi-label feature selection. In the upcoming section, we will discuss previous works in detail.

\subsection{Centralized Multi-label Feature Selection}
Multi-label feature selection methods can be divided into two groups based on how multi-label data is handled: problem transformation and algorithm adaptation. In problem transformation methods, the first step involves converting the multi-label dataset into a single-label one. Then, any state-of-the-art single-label FS method can be employed to select features effectively \cite{spolaor2013comparison, doquire2011feature, reyes2015scalable}. Binary relevance (BR) \cite{boutell2004learning}, label powerset (LP) \cite{tsoumakas2010random}, pruned problem transformation (PPT) \cite{read2008pruned}, and entropy-based label assignment (ELA) \cite{chen2007document} are a number of problem transformation methods that convert multi-label data into single-label format. For instance, in \cite{spolaor2013comparison}, the data is transformed using BR and LP techniques, followed by the utilization of Information Gain (IG) and ReliefF for feature selection. Similarly, Doquire et al. \cite{doquire2011feature} use PPT for problem transformation, and then employ a greedy forward feature selection method based on mutual information. However, these methods still have some drawbacks; for instance, BR does not take label correlations into account.

Algorithm adaptation approaches extend FS methods to directly handle multi-label datasets and effectively address those drawbacks \cite{hashemi2021efficient}. Until now, several algorithm adaptation methods have been proposed. For example, Lee et al. \cite{lee2015mutual} propose a FS method based on mutual information (D2F), which incorporates interaction information and utilizes conditional mutual information for assessing feature relevance. Additionally, Pairwise Multi-label Utility (PMU) method \cite{lee2013feature} is another approach that leverages the mutual information between a candidate feature and the label set, serving as a term for feature relevance. In \cite{lee2017scls}, a novel multi-label FS method named SCLS is proposed, which employs scalable feature relevance assessment to evaluate the relevance of candidate features.
\vspace{-2mm}
\subsection{Federated Feature Selection for Single-label Datasets}
There are a few number of federated feature selection (FFS) methods for single-label datasets in the literature \cite{mahanipour2023wrapper, zhang2023federated}. These methods are inspired by federated learning (FL) procedure and can be classified into two groups: vertical FFS \cite{li2023fedsdg} and horizontal FFS \cite{hu2022multi}. Vertical FFS involves clients' datasets containing instances with identical IDs but varying feature sets. Conversely, horizontal FFS is characterized by clients having distinct instances while utilizing identical feature sets. This paper presents the first horizontal FFS method designed for multi-label datasets.
\vspace{-5mm}
\section{Preliminaries}
\subsection{Pareto-based solutions} 
In multi-objective optimization problems (MOPs), unlike single-objective ones, conflicts between objectives result in a set of optimal solutions known as the Pareto optimal set, rather than a single optimal solution \cite{von2014survey}. We assume maximizing optimization while retaining generality for the concepts of Pareto optimality. The general MOP formula is as follows:

\begin{equation}\label{my_forth_eqn}
\begin{cases}
max \; O(g) = [o_1(g), o_2(g), \cdots, o_w(g)],\\
s.c.: g \in \Omega
\end{cases}
\end{equation}

\noindent where \(O(g) = [o_1(g), o_2(g), \cdots, o_w(g)]\) represents the objective vector, with \(w\,(w\geq 2)\) denoting the number of objective functions, and \(g=(g_1,\cdots,g_k)\) is the decision vector, where \(k\) is the number of decision variables. In MOP, there exist fundamental concepts, such as:

\begin{itemize}
\item Pareto dominance: For any two objective vectors \(u=(u_1,\cdots,u_w)\)and \(v=(v_1,\cdots,v_w)\), \(v\) is dominated by \(u\), denoted as \(u\succ v\), if and only if none of the elements in \(v\) exceed the corresponding elements in \(u\), and at least one element in \(u\) is strictly larger.

\begin{equation}\label{my_forth_eqn}
\small
\forall \,i \in (1,\cdots,w): u_i\geq v_i \ \mbox{\Large$\wedge$} \ \exists \, i\in(1,\cdots,w):u_i> v_i
\end{equation}

\item Pareto optimal set: A solution \(g \in \Omega\) that is not dominated by any other solution in \(\Omega\) belongs to the Pareto optimal set.
\vspace{-3mm}
\begin{equation}\label{my_forth_eqn}
\nexists \: g^{\prime} \in \Omega: \: g^{\prime} \succ g
\end{equation}

\item Pareto optimal front: The Pareto optimal front is defined as the projection of the Pareto optimal set onto the objective space.
\end{itemize}
\vspace{-3mm}
\subsection{Crowding distance}
The crowding distance of a solution is determined by the density of neighboring solutions around it. This metric is derived from the largest cube centered around the solution, excluding other solutions \cite{raquel2005effective}.
\vspace{-3mm}
\section{Proposed Method}
\subsection{System Overview}
We consider a two-tier setup, wherein horizontal FFS is executed. The first tier comprises various clients, including smart devices in contexts such as autonomous driving systems and healthcare systems, which gather multi-label data \(\mathcal U\). The second tier consists of an edge server \(\mathcal S\). Importantly, this approach can be scaled to handle a larger number of edge servers. We consider that there are a collection of \(M\) clients, denoted as \(C_m\), where \(m\) takes on values from 1 to \(M\), and an edge server denoted as \(e\). Here, \(M\) must be a minimum of 2, as having a single client would result in a centralized FS scenario. Each client's dataset \(\mathcal U=\mathbb R ^{N\times (D+L)}\) is represented as \(\mathcal U=\{(X_i, Y_i)\}_{i=1}^N\), where \(N\) is the number of instances. Each instance is associated with a \(D\)-dimensional feature vector \(x_i=(x_{i1}, x_{i2}, ..., x_{iD})\) and an \(L\)-dimensional label vector \(y_i=(y_{i1}, y_{i2}, ..., y_{iL})\). The label vector is a binary vector where \(y_{il}\) equals 1 only if the given instance obtains label \(Y_{l}\); otherwise, \(y_{il}\) equals 0. The structure of multi-label dataset at clients is depicted at Fig. 1.

\begin{figure}[t]
\center
\begin{tabular}{c c c c | c c c c}
\hline
\multicolumn{4}{c|} X & \multicolumn{4}{c} Y \\ \hline
$X_{1}$ & $X_{2}$ & $\cdots$ & $X_{D}$ & $Y_{1}$ & $Y_{2}$ & $\cdots$ & $Y_{L}$ \\ \hline
$x_{11}$ & $x_{12}$ & $\cdots$ & $x_{1D}$ & $y_{11}$ & $y_{12}$ & $\cdots$ & $y_{1L}$ \\
$x_{21}$ & $x_{22}$ & $\cdots$ & $x_{2D}$ & $y_{21}$ & $y_{22}$ & $\cdots$ & $y_{2L}$ \\
$\vdots$ & $\cdots$ & $\ddots$ & $\vdots$ & $\vdots$ & $\vdots$ & $\ddots$ & $\vdots$ \\
$x_{N1}$ & $x_{N2}$ & $\cdots$ & $x_{ND}$ & $y_{N1}$ & $y_{N2}$ & $\cdots$ & $y_{NL}$ \\\hline
\end{tabular}
\label{tab1}
\caption{Multi-Label Data Structure}
\end{figure}
\vspace{-2mm}
\subsection{Proposed Algorithm}
The proposed approach combines federated learning procedure with principles from information theory to select informative features from multi-label datasets across various clients. This technique can be referred to as FMLFS, which stands for Federated Multi-Label Feature Selection. In this method, our goal is to identify the most relevant features while minimizing redundancy. Therefore, we consider relevancy as the predictive power of features and redundancy of features as two objectives to convert the multi-label feature selection problem into a bi-objective optimization problem. Then, Pareto dominance and crowding distance concepts are employed to sort features. The FMLFS procedure comprises two phases: local phase in clients and global phase in the edge server. The overview of a distributed environment with multi-label datasets is depicted in Fig. 2.

\begin{figure}
\includegraphics[width=\linewidth]{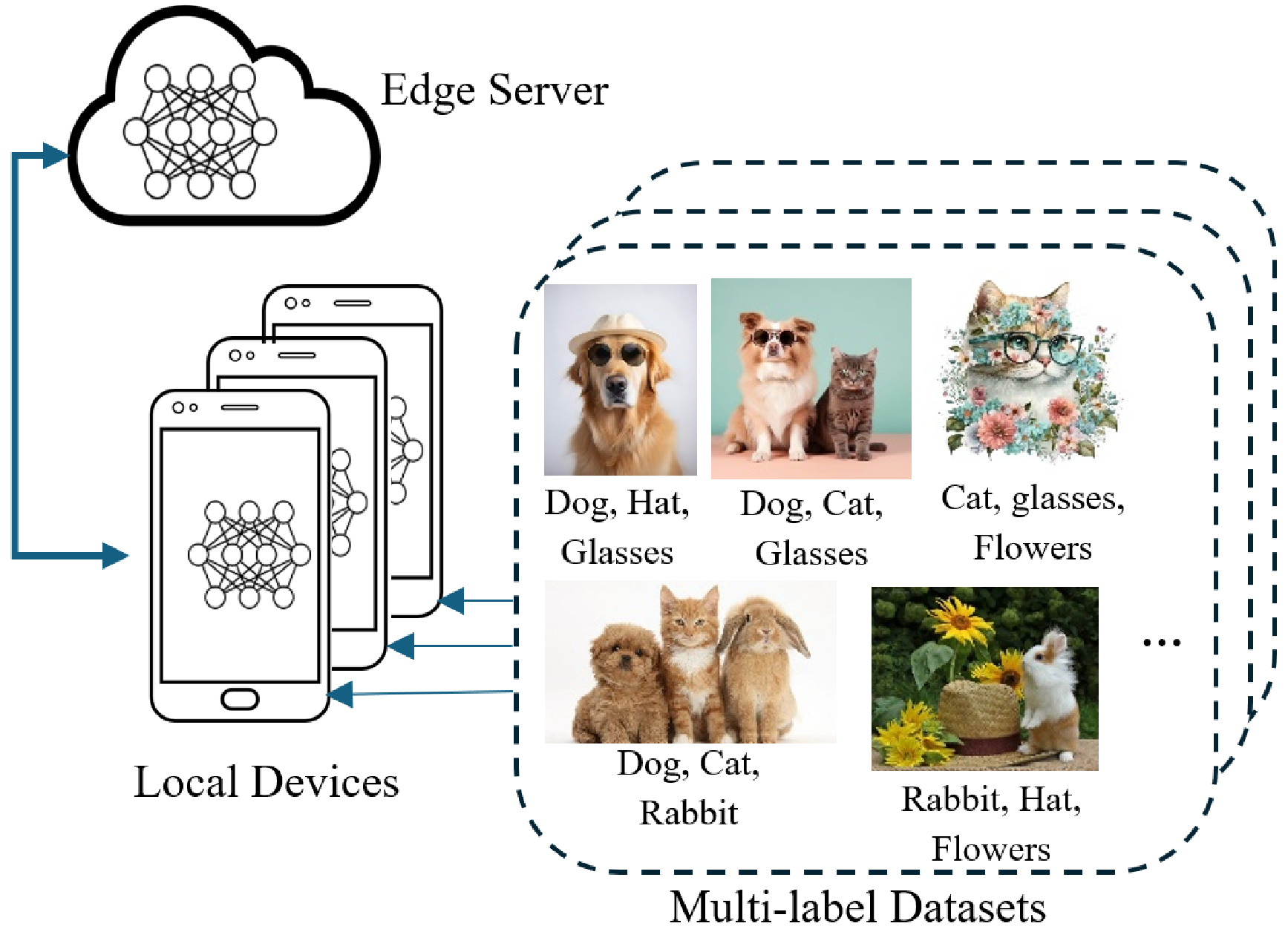}
\vspace{-5mm}
\caption{Overview of a distributed environment.}
\label{}
\end{figure}

\textbf{Local Phase:} In the local phase, each client uses its local dataset to calculate the mutual information between features and labels, determining their degree of relevance. Additionally, each client computes the correlation distance of features as the degree of redundancy by subtracting mutual information from joint entropy. Then, both the calculated mutual information and correlation distance measures are sent to the edge server for further processing.

Information entropy measures the uncertainty of random variables \cite{shannon2001mathematical}. The joint entropy between two features, such as \(X_{a}=(x_{1a}, x_{2a}, ..., x_{Na})\) and \(X_{b}=(x_{1b}, x_{2b}, ..., x_{Nb})\), is calculated as follows:

\begin{equation}\label{my_forth_eqn}
H(X_{a},X_{b})=H(X_{a}) + H(X_{a}|X_{b})
\end{equation}

\noindent where \(H(X_{a})\) represents the information entropy of the feature \(X_{a}\), and \(H(X_{a}|X_{b})\) denotes the conditional entropy of two features, defined as follows:
\vspace{-2mm}
\begin{equation}\label{my_forth_eqn}
H(X_{a})=-\sum_{i=1}^N p(x_{a}^i)log_2 p(x_{a}^i)
\end{equation}

\begin{equation}\label{my_sixth_eqn}
H(X_{a}|X_{b})=-\sum_{i=1}^N \sum_{j=1}^N p(x_{a}^i,x_{b}^j) log_2 p(x_{a}^i|x_{b}^j)
\end{equation}

\noindent \(p(x_{a}^i)\) represents the probability of the \(i\)-th value of feature \(X_{a}\), while \(p(x_{a}^i,x_{b}^j)\) denotes the joint probability of the \(i\)-th value of feature \(X_{a}\) and the \(j\)-th value of feature \(X_{b}\). Therefore, the joint entropy can be represented as follows: 
\vspace{-2mm}
\begin{equation}\label{my_forth_eqn}
H(X_{a},X_{b}) = -\sum_{i=1}^N\sum_{i=1}^N p(x_{a}^i,x_{b}^j) log_2 p(x_{a}^i,x_{b}^j)
\end{equation}

Mutual information quantifies the reduction in uncertainty of one random variable when another random variable is known, representing the amount of shared information between the variables. Consider \(X_{a}=(x_{1a}, x_{2a}, ..., x_{Na})\) as a feature and \(Y_{b}=(y_{1b}, y_{2b}, ..., y_{Nb})\) as a label in the dataset. The mutual information between the feature and label is defined as follows:

\begin{equation}\label{my_forth_eqn}
I(X_{a};Y_{b})=H(X_{a}) - H(X_{a}|Y_{b}) = H(Y_{b}) - H(Y_{b}|X_{a})
\end{equation}

\begin{equation}\label{my_forth_eqn}
I(X_{a};Y_{b})=H(X_{a}) + H(Y_{b}) - H(Y_{b},X_{a})
\end{equation}

Now, the correlation distance between two features can be defined as follows:

\begin{equation}\label{my_forth_eqn}
CD(X_{a},X_{b})=H(X_a,X_b)-I(X_a;X_b)
\end{equation}

\textbf{Global Phase:} On the edge server side, the computed mutual information (Eq. 11) and correlation distance (Eq. 12) matrices sent by the clients are aggregated to produce a global mutual information matrix and correlation distance matrix. These two properties are considered as two objectives in transforming the multi-label FS problem into a bi-objective optimization problem. Therefore, the optimal feature subset is defined as one that maximizes relevance while minimizing redundancy. 

\begin{equation}
MI=\begin{bmatrix}
I(X_1,Y_1) & I(X_1,Y_2) & \cdots & I(X_1,Y_L)\\
I(X_2,Y_1) & I(X_2,Y_2) & \cdots & I(X_2,Y_L)\\
\vdots & \vdots & \ddots & \vdots\\
I(X_D,Y_1) & I(X_D,Y_2) & \cdots & I(X_D,Y_L)
\end{bmatrix}
\end{equation}

\begin{equation}
\small
CD=\begin{bmatrix}
CD(X_1,X_1) & CD(X_1,X_2) & \cdots & CD(X_1,X_D)\\
CD(X_2,X_1) & CD(X_2,X_2) & \cdots & CD(X_2,Y_D)\\
\vdots & \vdots & \ddots & \vdots\\
CD(X_D,X_1) & CD(X_D,X_2) & \cdots & CD(X_D,X_D)
\end{bmatrix}
\end{equation}

Therefore, in defining the objective functions, the first objective is identified as maximizing the mutual information between each feature and the set of labels (\(O_1\)) \cite{hashemi2021efficient}: 

\begin{equation}\label{my_forth_eqn}
\begin{split}
&MAX(i) = max(MI(i,:)) , \quad i=1,2,...,D \\
&O_1 = [MAX(1), MAX(2), ..., MAX(D)]
\end{split}
\end{equation}

Next, the maximization of the correlation distance, defined as the difference between mutual information and joint entropy among features, is regarded as the second objective (\(O_2\)) to measure their redundancy:

\begin{equation}\label{my_forth_eqn}
\begin{split}
&A(i) = max(CD(i,:)) , \quad  i=1,2,...,D \\
&O_2 = [A(1), A(2), ..., A(D)]
\end{split}
\end{equation}

Then, a non-dominated sorting strategy is conducted using the Pareto dominance concept with these two objectives. Initially, each feature is assigned a Pareto number. Subsequently, the crowding distance or density of other features around each feature is calculated to arrange the features within the same front and with identical Pareto numbers. The crowding distance of each feature is computed within the bi-objective space. Next, the combination of the Pareto front number and the crowding distance is used to assign a final score to each feature. This score is calculated as follows \cite{hashemi2021efficient}:
\vspace{-3mm}
\begin{equation}\label{my_forth_eqn}
S = P + \frac{1}{(1+d)}
\end{equation}

\noindent where \(P\) denotes the Pareto front number, and \(d\) represents the crowding distance. A lower value of \(S\) indicates a better feature, as a lower Pareto front number and a larger crowding distance are considered preferable. The features can now be arranged based on their \(S\) values. The pseudocode of the FMLFS method is given in Algorithm 1 and 2. Also, the overview of the proposed method is demonstrated in Fig. 3.
\vspace{-2mm}
\begin{algorithm}
\caption{Pseudocode of the edge server side}\label{alg:cap}
\begin{algorithmic}[1]
\renewcommand{\algorithmicrequire}{\textbf{Input:}}
\renewcommand{\algorithmicensure}{\textbf{Output:}}
\Require M (number of clients), clients' \(MI\) matrices, clients' correlation distance matrices (\(CD\))
\Ensure Ranking of features based on \(S\)
\newline
\State Executing \textbf{Algorithm 2} in clients

\noindent \# Aggregation step
\State \(MI^{\prime} = (MI_1 + MI_2 + ... +MI_M)/M \) 
\State \(CD^{\prime} = (CD_1 + CD_2 +...+ CD_M)/M\)

\noindent\# Calculation of Objective functions
\State \(O_1 = \max(MI^{\prime},1)\)
\State \(O_2 = \max(CD^{\prime},1)\)

\noindent\# Feature sorting
\State Performing the non-dominated sorting algorithm in the bi-objective domain 
\State Assigning Pareto front number \(P\) to the features
\State Calculating the crowding distance of features \(d\)
\State \(S = P + \frac{1}{(1+d)}\) 
\State Sorting features based on \(S\) in ascending order

\end{algorithmic}
\end{algorithm}

\begin{algorithm}
\caption{Pseudocode of the client side}\label{alg:cap}
\begin{algorithmic}[1]
\renewcommand{\algorithmicrequire}{\textbf{Input:}}
\renewcommand{\algorithmicensure}{\textbf{Output:}}
\Require Local dataset of each client
\Ensure Mutual information matrix (\(MI\)), and correlation distance matrix (\(CD\))
\newline

\noindent\# Calculating Mutual information between features and labels
\For{\texttt{a=1:D}}
    \State \For{\texttt{b=1:L}}
        \State \texttt{\(I(X_a;Y_b)\)}
    \EndFor
\EndFor

\noindent\# Calculating correlation distance between features
\For{\texttt{a=1:D}}
    \State \For{\texttt{b=1:D}}
        \State \texttt{\(CD(X_a,X_b)=H(X_a,X_b)-I(X_a;X_b)\)}
    \EndFor
\EndFor
\State return the MI and CD matrices
\end{algorithmic}
\end{algorithm}

\begin{figure}
\includegraphics[width=\linewidth]{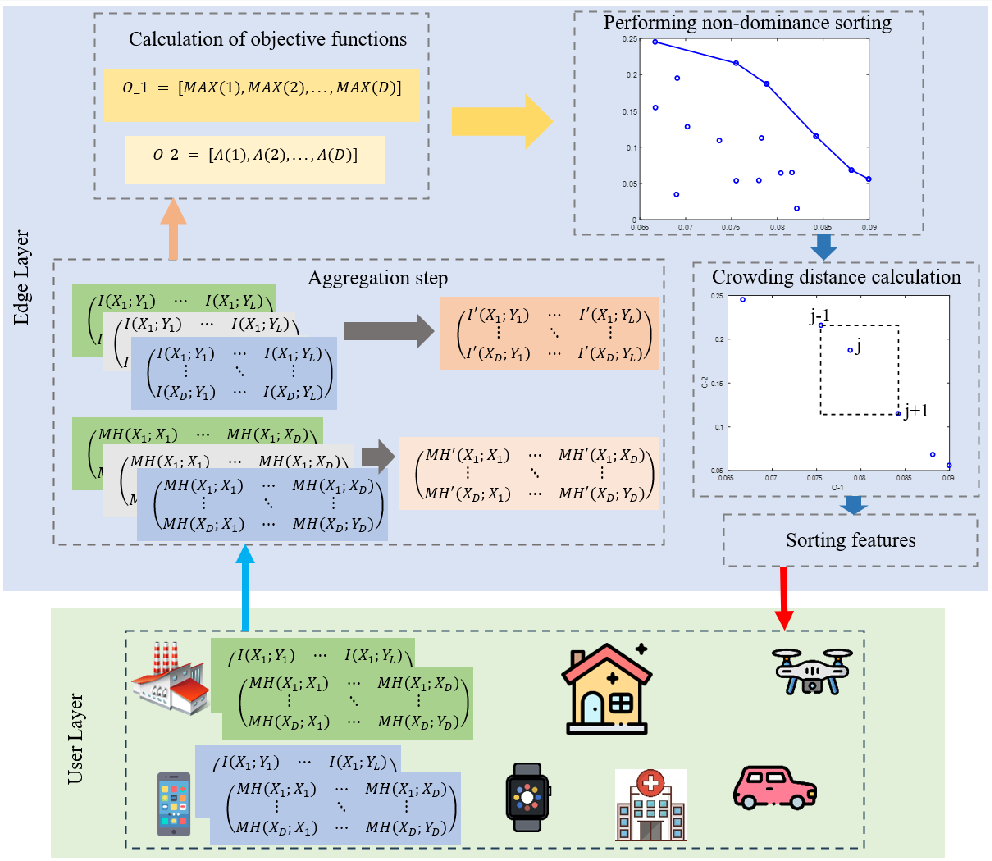}
\vspace{-5mm}
\caption{Overview of the FMLFS algorithm.}
\label{}
\end{figure}
\vspace{-5mm}
\section{Experimental Results}
In this section, we evaluate the proposed method using two scenarios. In the first scenario, clients rank features by FMLFS and select the desired number of features. Subsequently, the reduced-size datasets are transmitted to the edge server to be utilized in a centralized multi-label learning algorithm. In the second scenario, after employing FMLFS to rank features, a vanilla federated learning algorithm is utilized as a multi-label classifier.
\vspace{-2mm}
\subsection{Datasets}
The proposed method's performance is evaluated against five similar methods in the literature using three real-world datasets from the Mulan\footnote[1]{https://mulan.sourceforge.net/datasets.html} repository. The datasets are selected from diverse domains (Biology, Image, and Audio), each varying in the number of instances, features, and labels. The characteristics of these datasets are presented in Table I.

\begin{table}[htbp]

\caption{Details of the multi-label datasets}
\vspace{-4mm}
\begin{center}
\resizebox{\columnwidth}{!}{%
\begin{tabular}{c|c|c|c|c}
\hline
Dataset & Instances & Features & Labels & Domain \\
\hline
Dataset & Instances & Features & Labels & Domain \\
\hline
Yeast & 2417 & 103 & 14 & Biology\\

Scene & 2407 & 294 & 6 & Image\\

Birds & 645 & 260 & 19 & Audio\\
\hline
\end{tabular}
}
\label{tab1}
\end{center}
\vspace{-5mm}
\end{table}
\vspace{-2mm}
\subsection{Evaluation Measures}

Accuracy, F-measure, hamming loss, ranking loss, average precision and coverage are the metrics employed to evaluate the performance of FMLFS and other comparative methods. Let (\(\mathcal T=\{(x_i, y_i)\}_{i=1}^n\)) denote a test set, where \(y_i\) and \(z_i\) represent the actual label set and the predicted label set for \(x_i\) respectively. Now, let's define the metrics as follows \cite{tarekegn2021review}:

\begin{itemize}
\item Accuracy: It represents the proportion of correctly predicted labels relative to all predicted and actual labels.
\vspace{-2mm}
\begin{equation}\label{my_forth_eqn}
Accuracy = \frac{1}{n} \sum_{i=1}^n \frac{|y_i \cap z_i|}{|y_i \cup z_i|}
\end{equation}

\item F-measure: It is a harmonic mean of precision and recall. It is a weighted measure indicating the number of relevant labels predicted and the proportion of predicted labels that are relevant.
\vspace{-2mm}
\begin{equation}\label{my_forth_eqn}
\begin{split}
&F-measure = 2\times \frac{Precision\times Recall}{Precision + Recall} \\
&Precision = \frac{1}{n} \sum_{i=1}^n \frac{|y_i \cap z_i|}{|z_i|}\\
&Recall = \frac{1}{n} \sum_{i=1}^n \frac{|y_i \cap z_i|}{|y_i|}
\end{split}
\end{equation}

\item Hamming Loss (HL): It is calculated by determining the symmetric difference between the actual and predicted labels and then dividing it by the total number of labels. A smaller HL value indicates better performance.
\vspace{-2mm}
\begin{equation}\label{my_forth_eqn}
HL = \frac{1}{n} \sum_{i=1}^n \frac{|y_i \triangle z_i|}{|L|}
\end{equation}

\item Ranking Loss (RL): It calculates the frequency of relevant labels being ranked lower than non-relevant labels. Better performance is indicated by a smaller RL value.
\vspace{-2mm}
\begin{equation}\label{my_forth_eqn}
\begin{split}
&RL = \frac{1}{n} \sum_{i=1}^n \frac{1}{|y_i||\Bar{y_i}|} |{(\lambda_a,\lambda_b):rank(\lambda_a)> rank(\lambda_b),}\\
& {(\lambda_a,\lambda_b)\in y_i \times \Bar{y_i}}|
\end{split}
\end{equation}
\noindent where \(\Bar{y_i}\) is the complement set of \(y_i\).

\item Avg-Precision: It measures the average fraction of relevant labels ranked above a specific label.
\vspace{-2mm}
\begin{equation}\label{my_forth_eqn}
\begin{split}
\small
\medmath {Avg-precision = \frac{1}{n} \sum_{i=1}^n \frac{1}{|y_i|}
\sum_{\lambda \in y_i} \frac{|{\lambda^{\prime} \in y_i : rank(\lambda^{\prime}) \leq rank(\lambda)}|}{rank(\lambda)}}
\end{split}
\end{equation}

\item Coverage: It denotes the number of steps a learning algorithm requires to cover all the true labels of an instance. The better the performance is indicated by a smaller coverage value.
\vspace{-3mm}
\begin{equation}\label{my_forth_eqn}
Coverage = \frac{1}{n} \sum_{i=1}^n \max_{\lambda \in y_i} (rank(\lambda)) -1
\end{equation}
\end{itemize}
\vspace{-2mm}
\subsection{Parameter setting}
\vspace{-1mm}
In this work, two scenarios are considered to evaluate the performance of the proposed method. As mentioned before, in the first scenario, after ranking and selecting the desired number of features, local datasets are transmitted to the edge server. Here, ML-kNN \cite{zhang2007ml} with \(k=10\) is used as a classifier, representing one of the most commonly utilized learning algorithms in centralized multi-label classification. In the second scenario, the vanilla federated learning algorithm with multi layer perceptron (MLP) is employed after ranking and selecting the desired number of features. 

Throughout our experiments, we utilize 10 clients, consistent with other single-label federated feature selection methods in the literature. Additionally, the data demonstrates non-independent and non-identically distributed (Non-IID) characteristics across the clients.
\vspace{-2mm}
\begin{figure}[htbp]
  \centering
  \begin{tabular}[c]{cc}
    \begin{subfigure}{0.22\textwidth}
      \includegraphics[width=\textwidth]{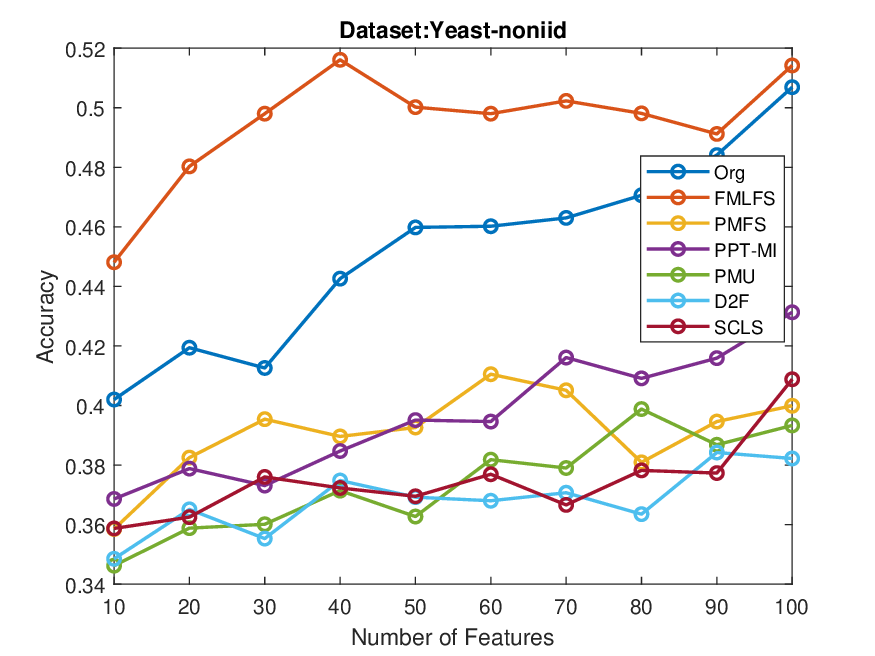}
      \caption{$Accuracy$}
    \end{subfigure}&
    \begin{subfigure}{0.22\textwidth}
      \includegraphics[width=\textwidth]{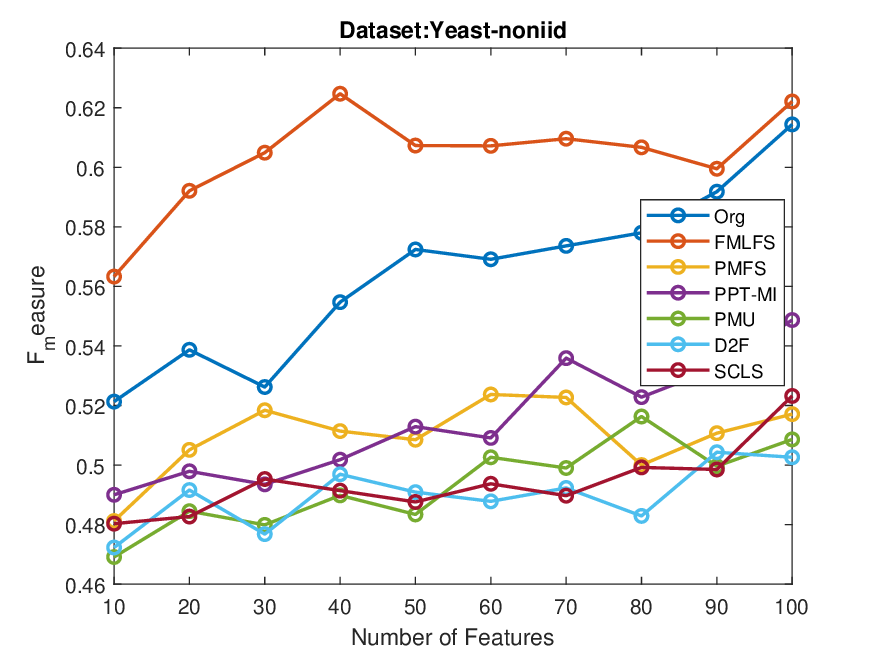}
      \caption{$F-measure$}
    \end{subfigure}\\
    \begin{subfigure}{0.22\textwidth}
      \includegraphics[width=\textwidth]{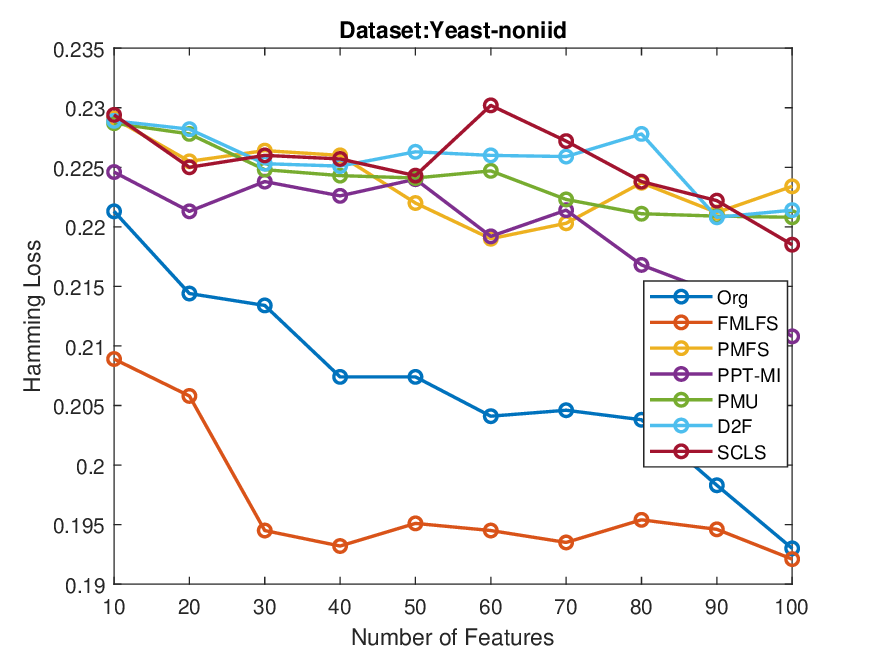}
      \caption{$Hamming Loss$}
    \end{subfigure}&
    \begin{subfigure}{0.22\textwidth}
      \includegraphics[width=\textwidth]{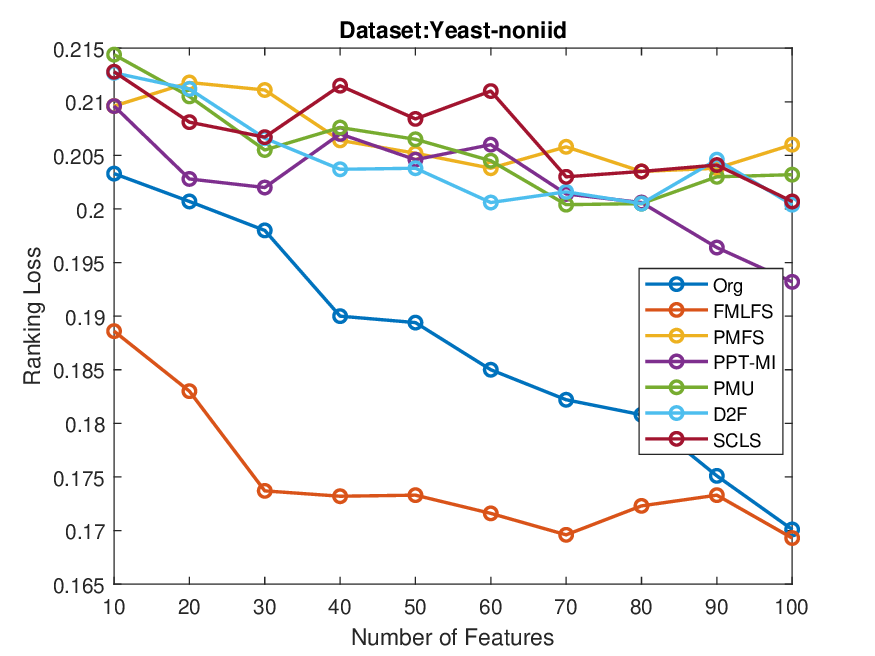}
      \caption{$Ranking Loss$}
    \end{subfigure}\\
    \begin{subfigure}{0.22\textwidth}
      \includegraphics[width=\textwidth]{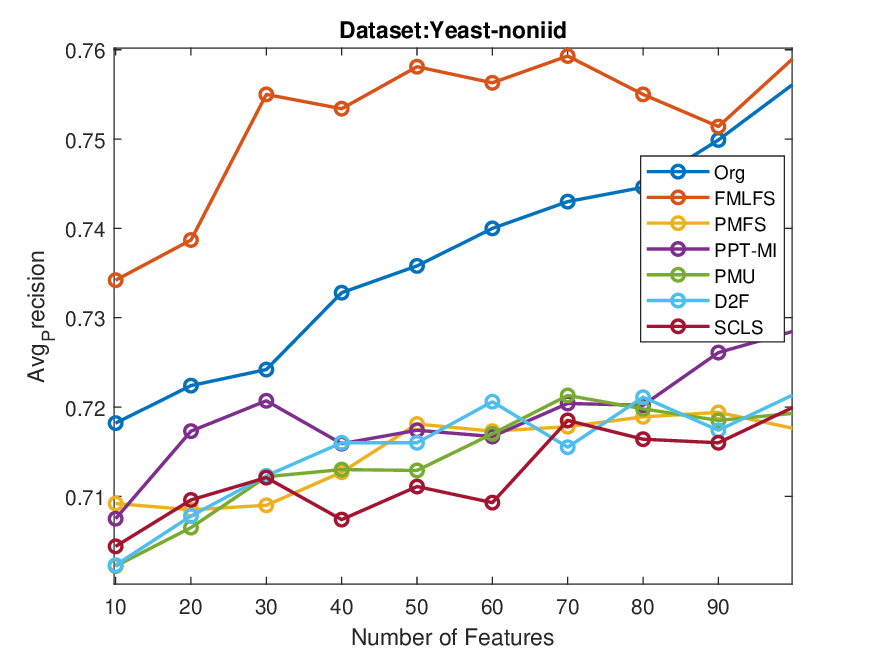}
      \caption{$Avg Precision$}
    \end{subfigure}&
    \begin{subfigure}{0.22\textwidth}
      \includegraphics[width=\textwidth]{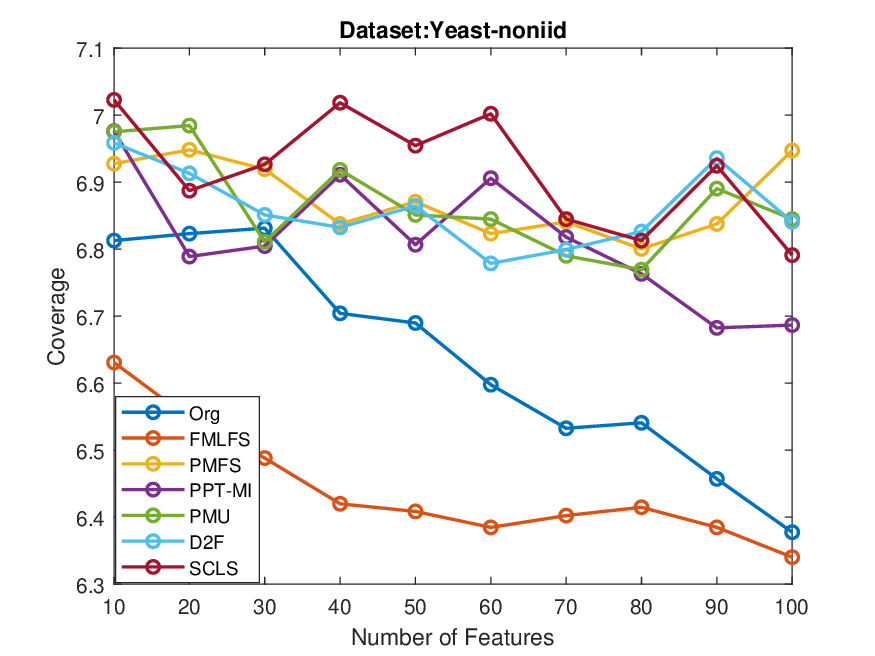}
      \caption{$Coverage$}
    \end{subfigure}
  \end{tabular}
  \vspace{-3mm}
  \caption{Results for Yeast non-iid dataset with ML-kNN.}\label{fig:animals}
\end{figure}

\begin{figure}[htbp]
  \centering
  \begin{tabular}[c]{cc}
    \begin{subfigure}{0.22\textwidth}
      \includegraphics[width=\textwidth]{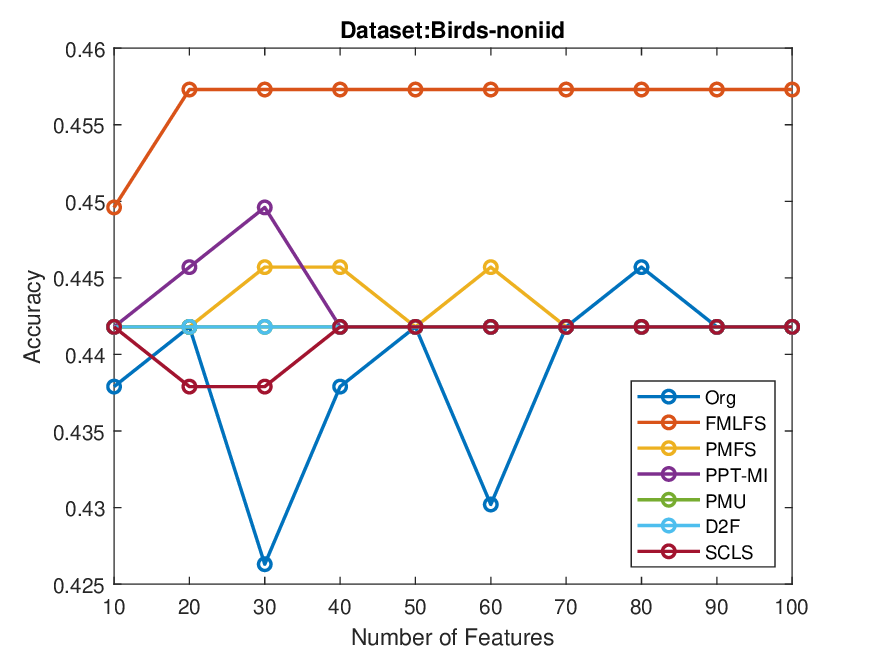}
      \caption{$Accuracy$}
    \end{subfigure}&
    \begin{subfigure}{0.22\textwidth}
      \includegraphics[width=\textwidth]{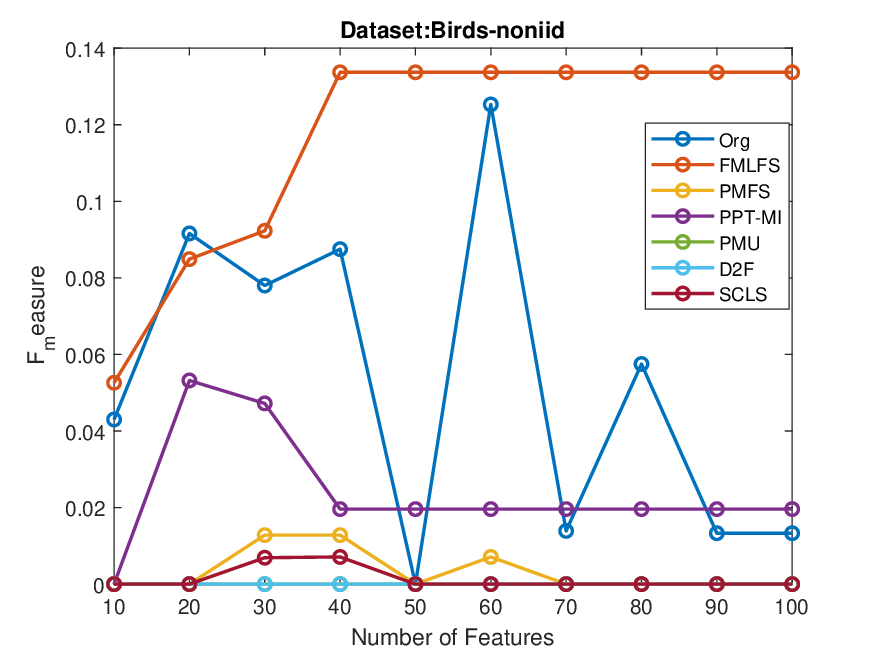}
      \caption{$F-measure$}
    \end{subfigure}\\
    \begin{subfigure}{0.22\textwidth}
      \includegraphics[width=\textwidth]{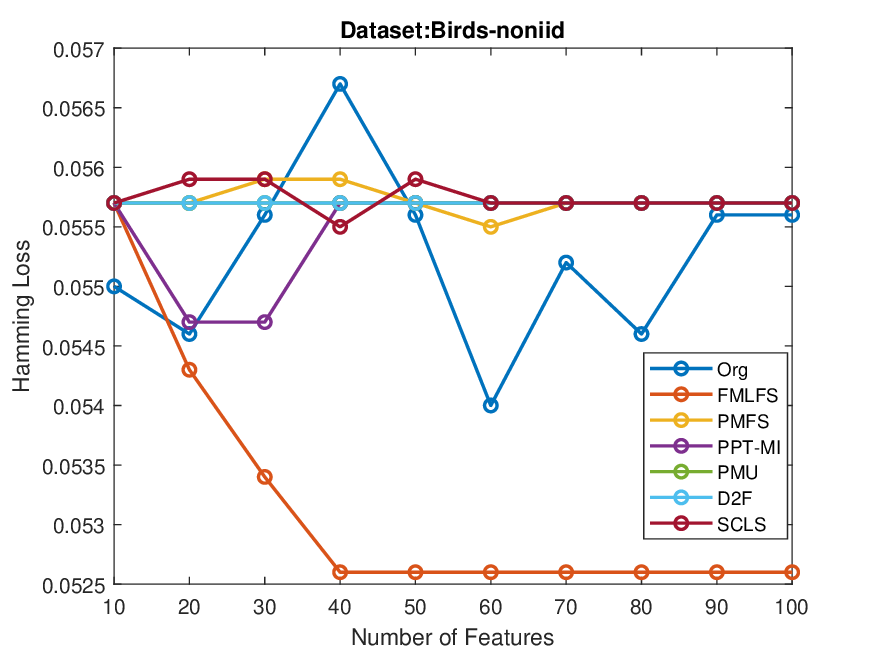}
      \caption{$Hamming Loss$}
    \end{subfigure}&
    \begin{subfigure}{0.22\textwidth}
      \includegraphics[width=\textwidth]{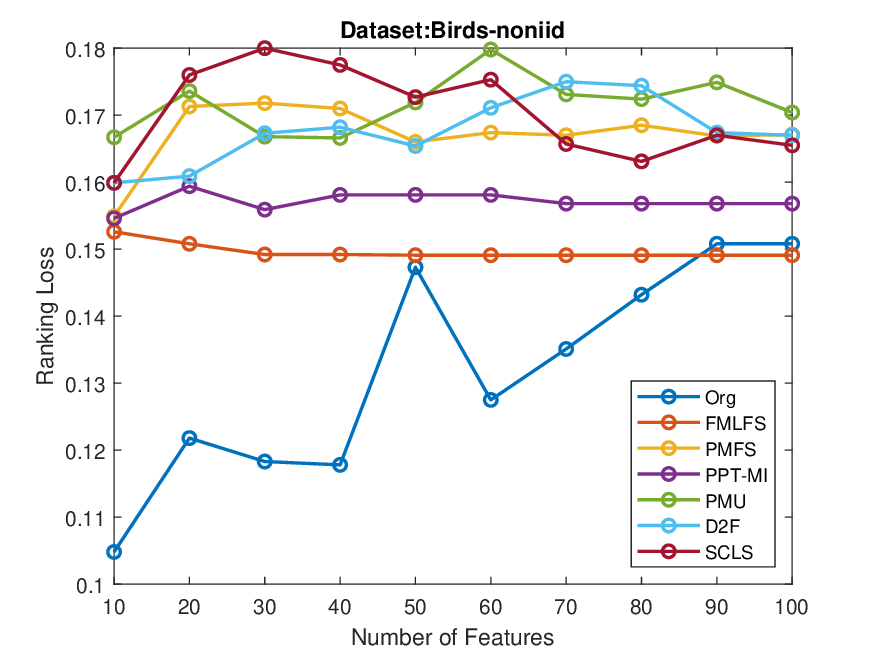}
      \caption{$Ranking Loss$}
    \end{subfigure}\\
    \begin{subfigure}{0.22\textwidth}
      \includegraphics[width=\textwidth]{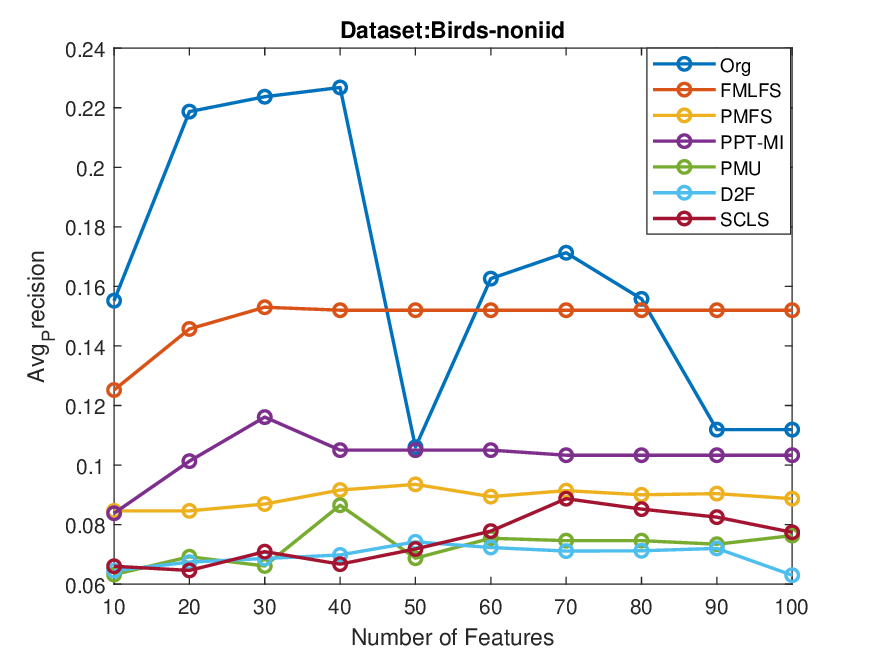}
      \caption{$Avg Precision$}
    \end{subfigure}&
    \begin{subfigure}{0.22\textwidth}
      \includegraphics[width=\textwidth]{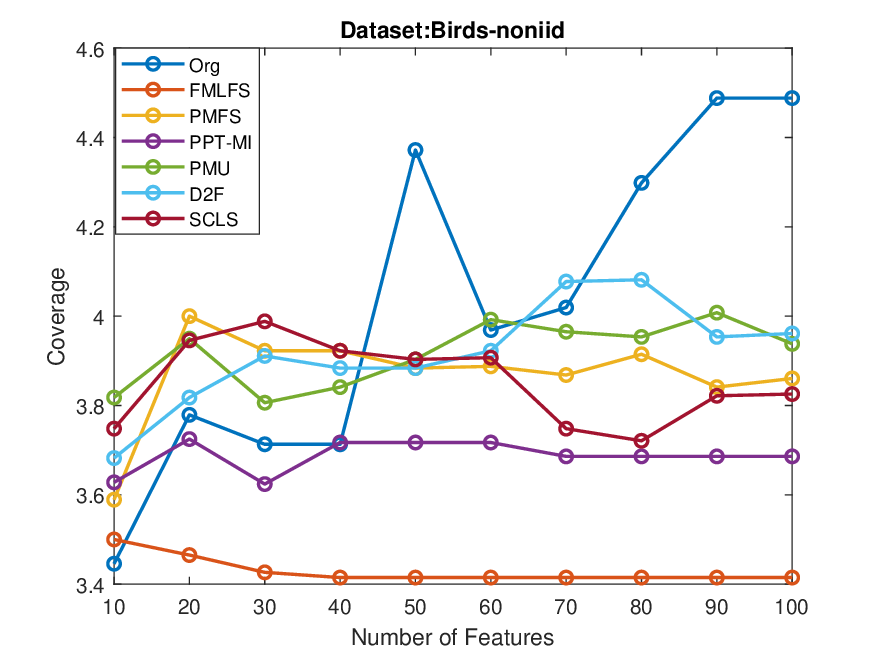}
      \caption{$Coverage$}
    \end{subfigure}
  \end{tabular}
  \vspace{-3mm}
  \caption{Results for Birds non-iid dataset with ML-kNN.}\label{fig:animals}
\end{figure}

\begin{figure}[htbp]
  \centering
  \begin{tabular}[c]{cc}
    \begin{subfigure}{0.22\textwidth}
      \includegraphics[width=\textwidth]{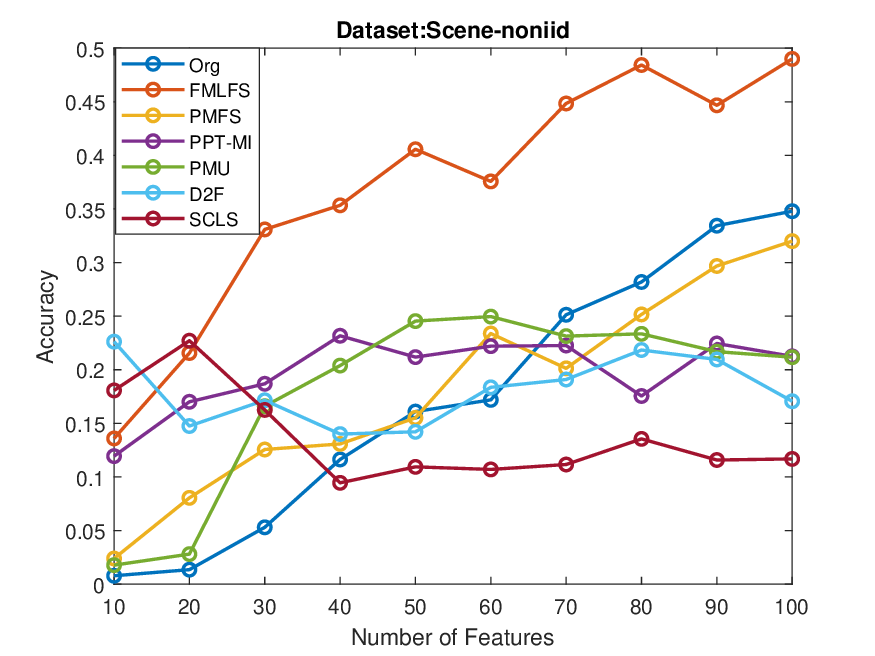}
      \caption{$Accuracy$}
    \end{subfigure}&
    \begin{subfigure}{0.22\textwidth}
      \includegraphics[width=\textwidth]{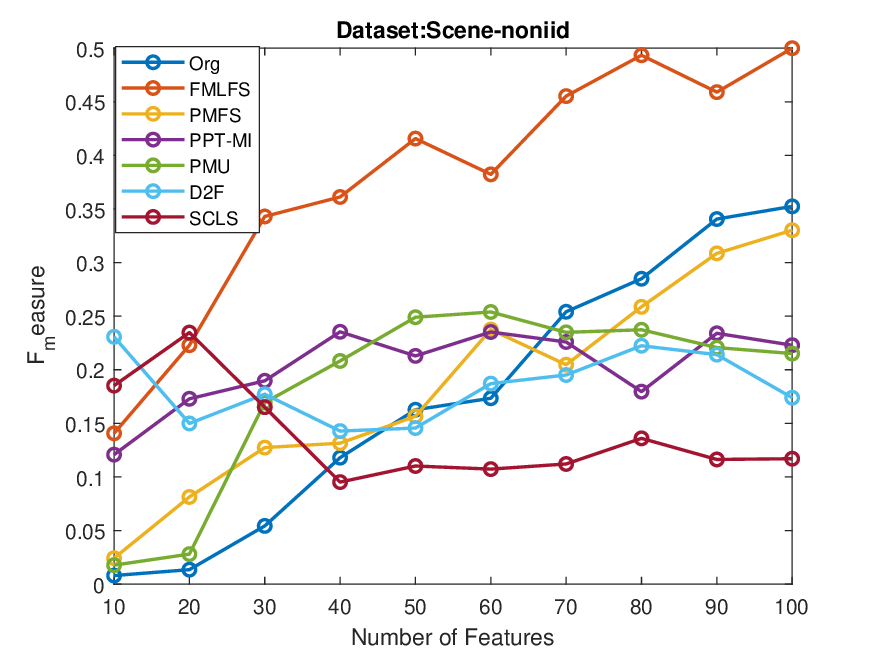}
      \caption{$F-measure$}
    \end{subfigure}\\
    \begin{subfigure}{0.22\textwidth}
      \includegraphics[width=\textwidth]{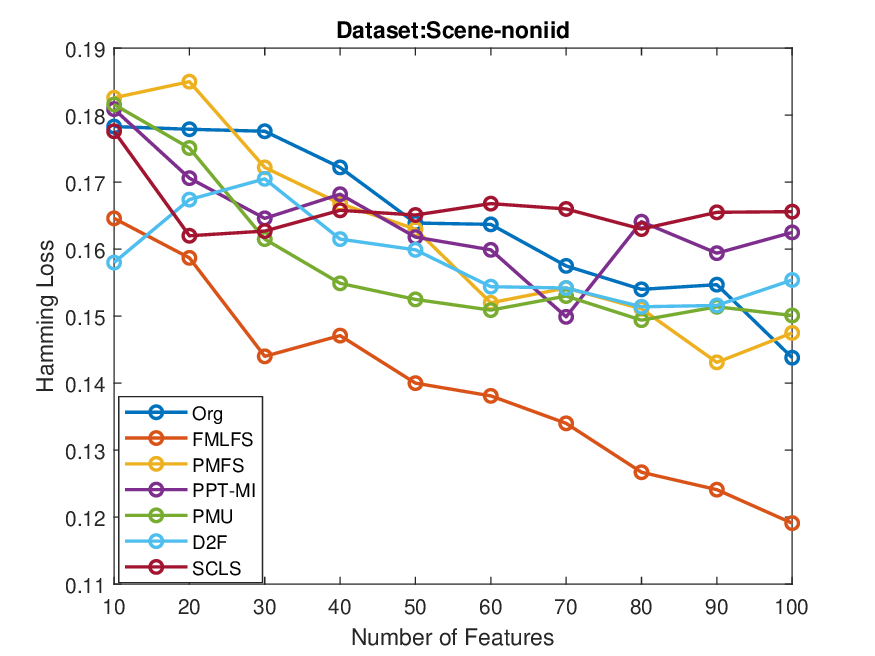}
      \caption{$Hamming Loss$}
    \end{subfigure}&
    \begin{subfigure}{0.22\textwidth}
      \includegraphics[width=\textwidth]{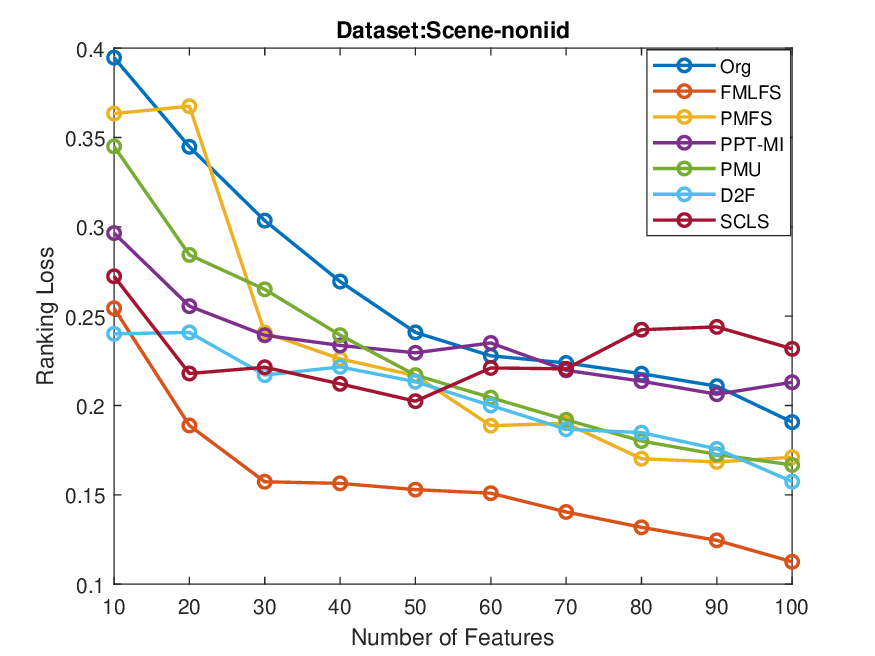}
      \caption{$Ranking Loss$}
    \end{subfigure}\\
    \begin{subfigure}{0.22\textwidth}
      \includegraphics[width=\textwidth]{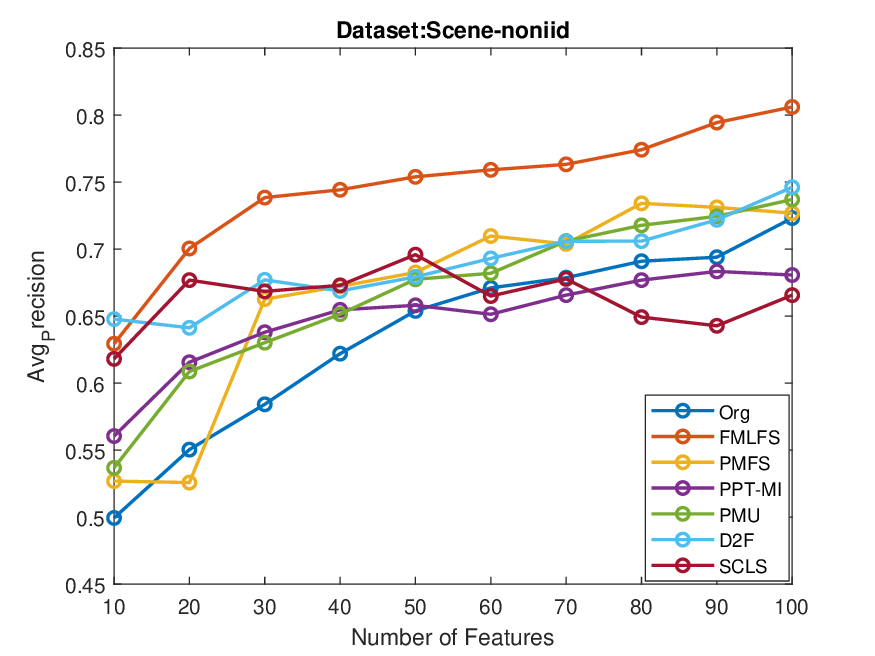}
      \caption{$Avg Precision$}
    \end{subfigure}&
    \begin{subfigure}{0.22\textwidth}
      \includegraphics[width=\textwidth]{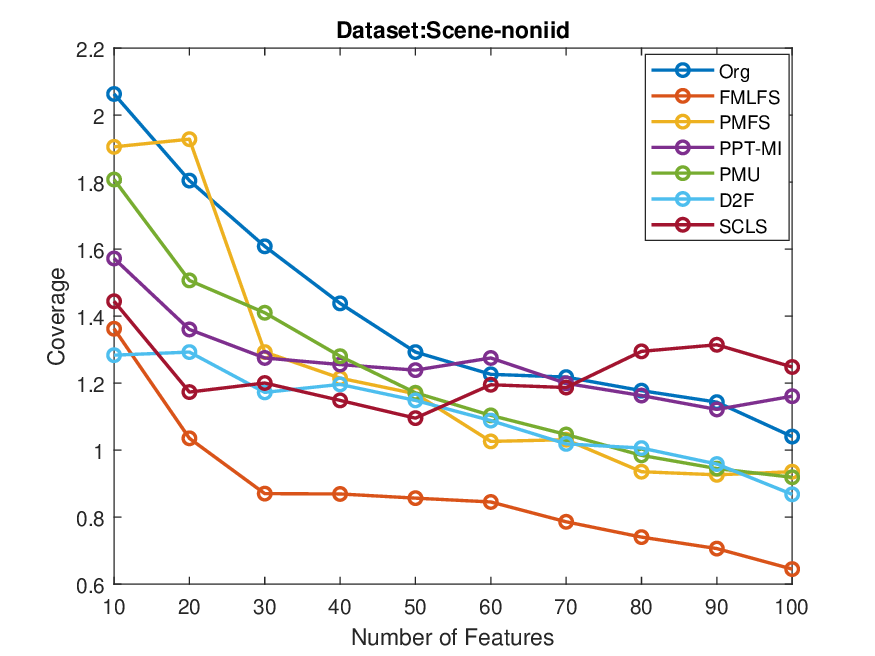}
      \caption{$Coverage$}
    \end{subfigure}
  \end{tabular}
  \vspace{-3mm}
  \caption{Results for Scene non-iid dataset with ML-kNN.}\label{fig:animals}
\end{figure}
\subsection{Results and Analysis}
\vspace{-1mm}
In this study, we consider three metrics—performance, computational complexity, and communication cost—to evaluate the effectiveness of the proposed method compared to five other methods. Performance is evaluated using six metrics: accuracy, F-measure, hamming loss, ranking loss, average precision, and coverage. Computational complexity is assessed by examining the time complexity of each algorithm, while communication cost is determined based on the size of the dataset.

In the first scenario, we compare the proposed method, the first federated multi-label feature selection method in the literature, with five other centralized multi-label feature selection methods including: PMFS \cite{hashemi2021efficient}, PPT-MI \cite{doquire2011feature}, PMU \cite{lee2013feature}, D2F \cite{lee2015mutual}, and SCLS \cite{lee2017scls}. Our proposed method (FMLFS) ranks features across all clients in a federated manner. In contrast, the other methods operate independently within each client, with no communication between clients or between clients and the server. After ranking features within each client, the desired number of features is selected. Then, the reduced datasets are transmitted from clients to the edge server to feed into the centralized ML-kNN classifier. The results of this scenario are presented in Fig. 4 to 6. 

In the second scenario, the proposed method is also compared with the five existing methods. The main difference compared to the first scenario is that, after feature ranking, the federated learning model is modified to function as a multi-label classifier. The findings of this scenario are depicted in Fig. 7 to 9.

\textbf{Discussion:} In the first scenario, for both the Yeast and Scene datasets, the proposed method demonstrates superior performance across all six evaluation metrics compared to the five other methods and the original dataset without FS. For instance, in the Yeast dataset, FMLFS achieves an accuracy of 0.48 with just 20 features, which is comparable to the performance of the classifier using 90 features without FS on the cloud server. It's worth noting that the cloud server is at least 10 times farther away than the edge server. Therefore, FMLFS can effectively reduce communication costs while simultaneously improving the learning algorithm's performance, offering a good trade-off between performance and communication cost. Additionally, in the Yeast dataset, FMLFS demonstrates better results with just 10 features across all evaluation metrics compared to the five other FS methods with 100 features. Moreover, in the Birds dataset, FMLFS outperforms all other methods, although the original dataset yields better results in terms of ranking loss and average precision.

In the second scenario, it is evident that the performance of FMLFS with 10 features in the Yeast and Scene datasets surpasses that of other methods using 100 features. This underscores the ability of the proposed method to provide a good trade-off between performance and computational complexity of the learning algorithm. Furthermore, in the Birds dataset, it achieves comparable or even better performance compared to the original dataset without FS, particularly in terms of average precision, coverage, and ranking loss.

\textbf{Time complexity analysis:} Here, we present the time complexity of FMLFS and the five other compared methods (PMFS, PPT-MI, PMU, D2F, and SCLS) on the client side, which is more important due to the constrained computational capabilities of end-user devices. Let \(N\), \(D\), and \(L\) represent the number of instances, the number of features and the number of labels, respectively. The time complexity of mutual information and joint entropy is \(O(N)\) because accessing all instances is required for probability calculation \cite{hu2022feature}. Time complexity of PMFS method is \(O(D^3 + D^2N + DNL)\). PPT-MI computes mutual information between features and each label, thus resulting in a time complexity of \(O(ND)\). If we denote the number of selected features in PMU as \(k\), its time complexity can be expressed as \(O(NDL + kNDL + NDL^2)\). The time complexity of D2F is \(O(NDL + kNDL)\), where the feature relevance and feature redundancy terms have time complexities of \(O(NDL)\) and \(O(kNDL)\) respectively. Also, the time complexity of SCLS is \(O(NDL + kND)\). The time complexities of the feature relevance and feature redundancy terms in our proposed method are \(O(NDL)\) and \(O(ND^2)\) respectively. Therefore, the overall time complexity is \(O(NDL + ND^2)\), which is the same as or even less than that of other compared methods.
\vspace{-2mm}

\begin{figure}[htbp]
  \centering
  \begin{tabular}[c]{cc}
    \begin{subfigure}{0.22\textwidth}
      \includegraphics[width=\textwidth]{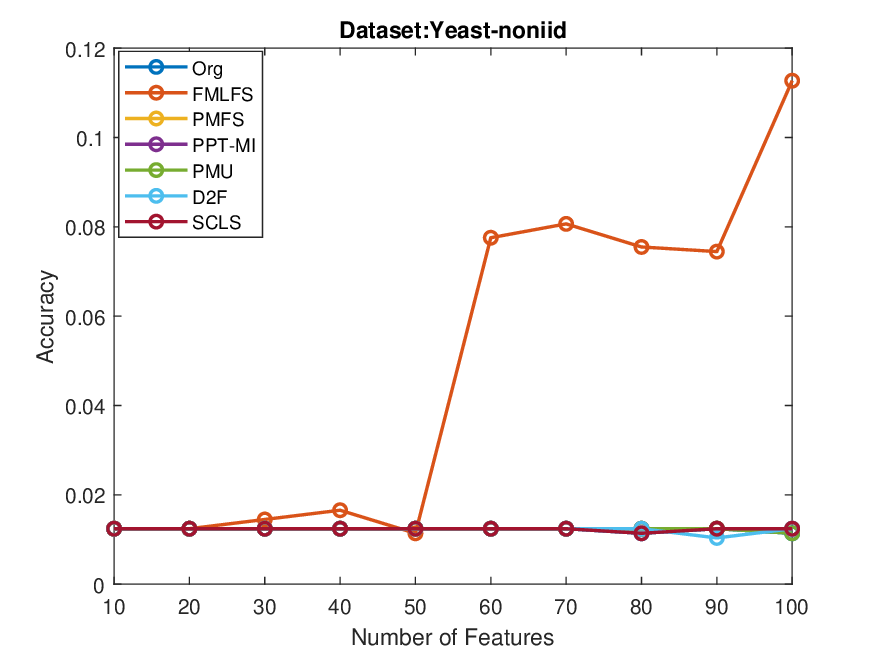}
      \caption{$Accuracy$}
    \end{subfigure}&
    \begin{subfigure}{0.22\textwidth}
      \includegraphics[width=\textwidth]{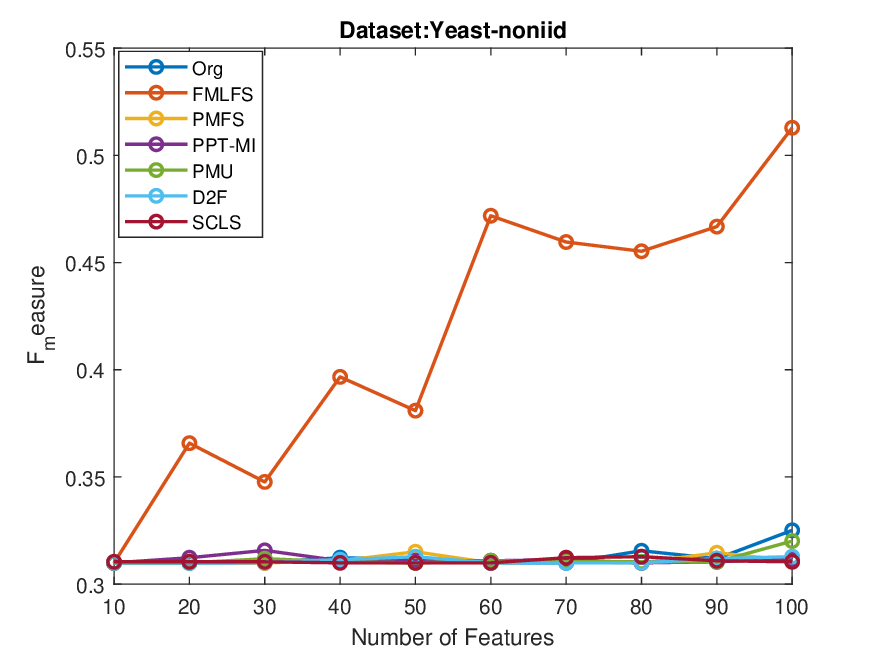}
      \caption{$F-measure$}
    \end{subfigure}\\
    \begin{subfigure}{0.22\textwidth}
      \includegraphics[width=\textwidth]{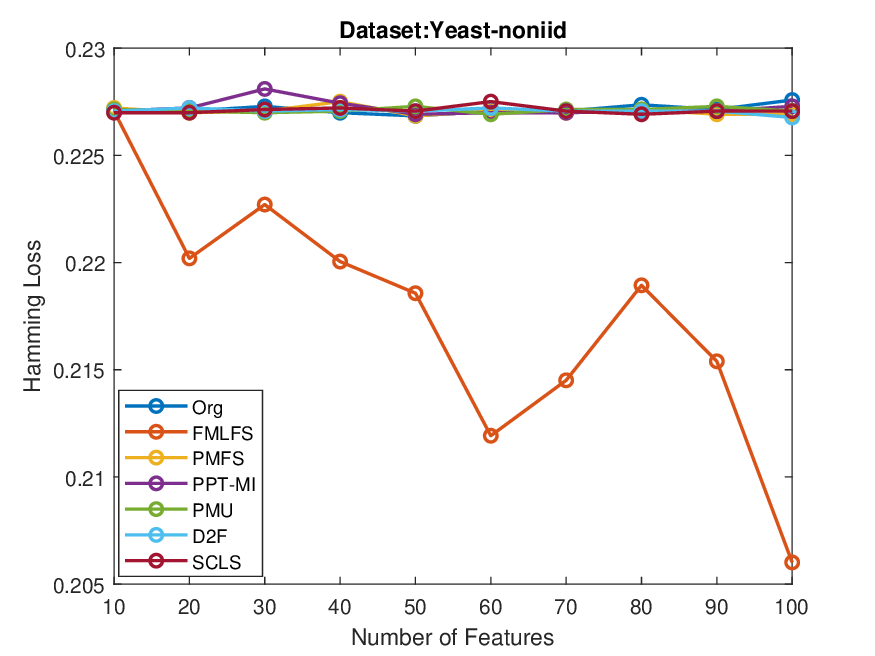}
      \caption{$Hamming Loss$}
    \end{subfigure}&
    \begin{subfigure}{0.22\textwidth}
      \includegraphics[width=\textwidth]{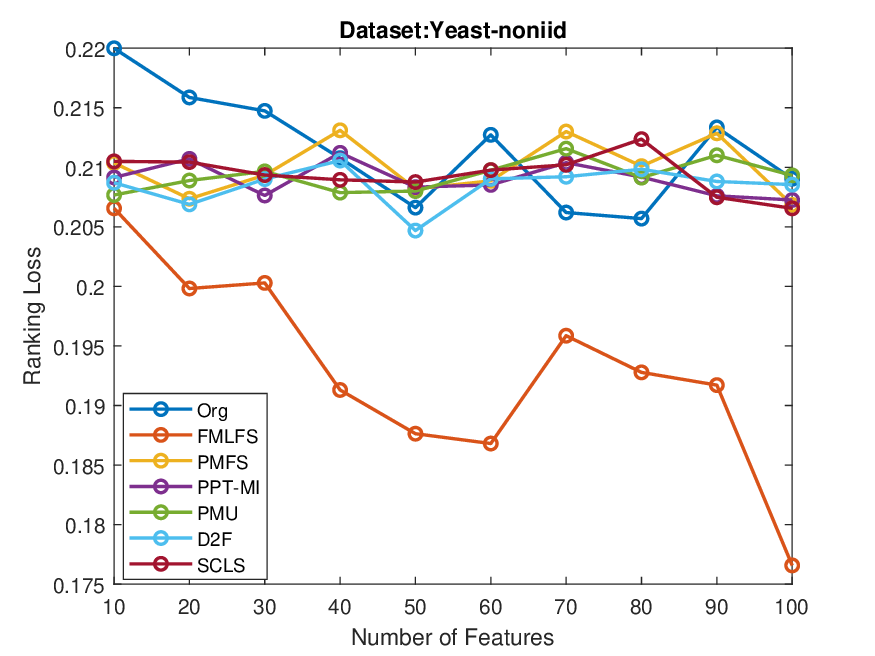}
      \caption{$Ranking Loss$}
    \end{subfigure}\\
    \begin{subfigure}{0.22\textwidth}
      \includegraphics[width=\textwidth]{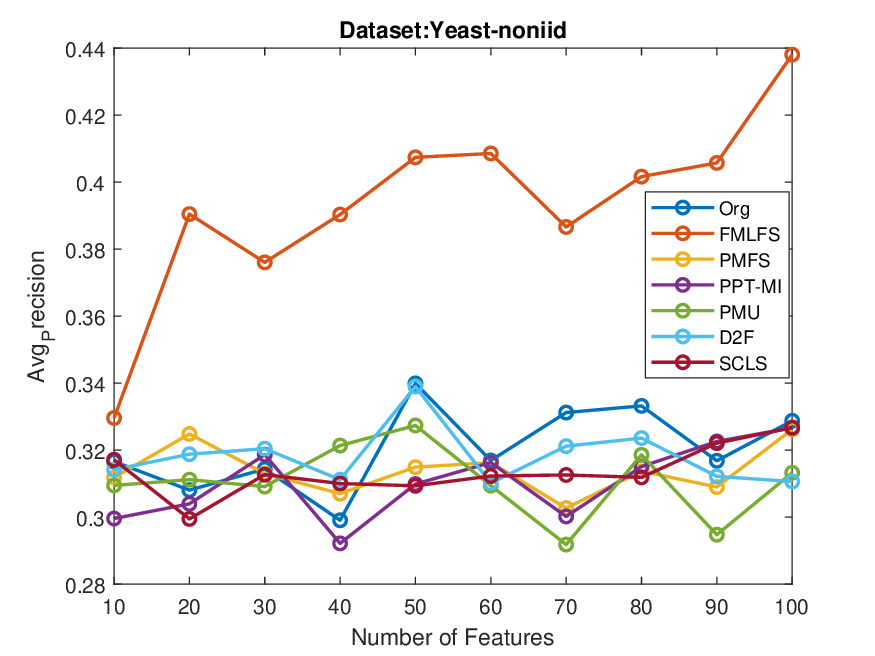}
      \caption{$Avg Precision$}
    \end{subfigure}&
    \begin{subfigure}{0.22\textwidth}
      \includegraphics[width=\textwidth]{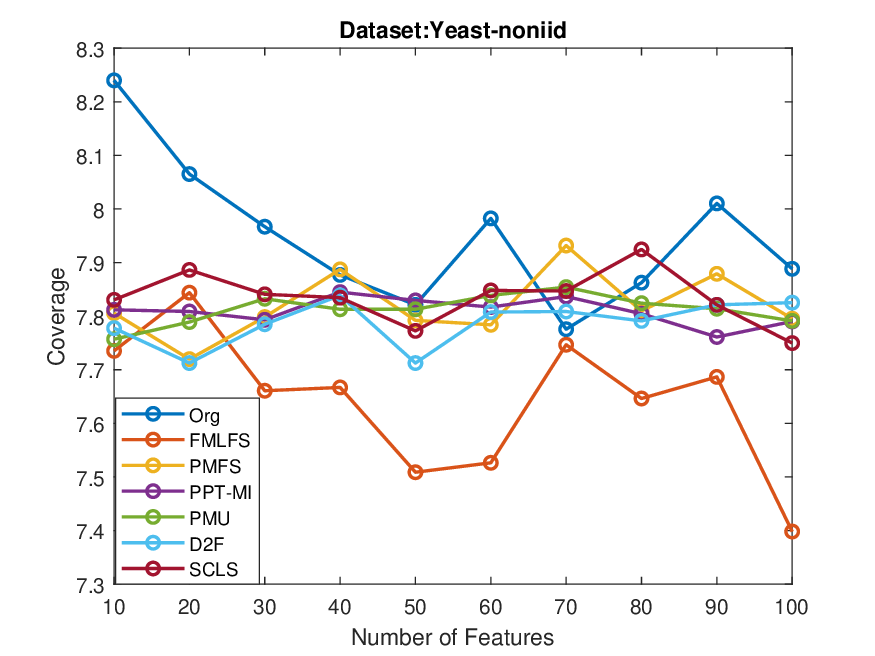}
      \caption{$Coverage$}
    \end{subfigure}
  \end{tabular}
  \vspace{-3mm}
  \caption{Results for Yeast non-iid dataset with FL.}\label{fig:animals}
\end{figure}

\begin{figure}[htbp]
  \centering
  \begin{tabular}[c]{cc}
    \begin{subfigure}{0.22\textwidth}
      \includegraphics[width=\textwidth]{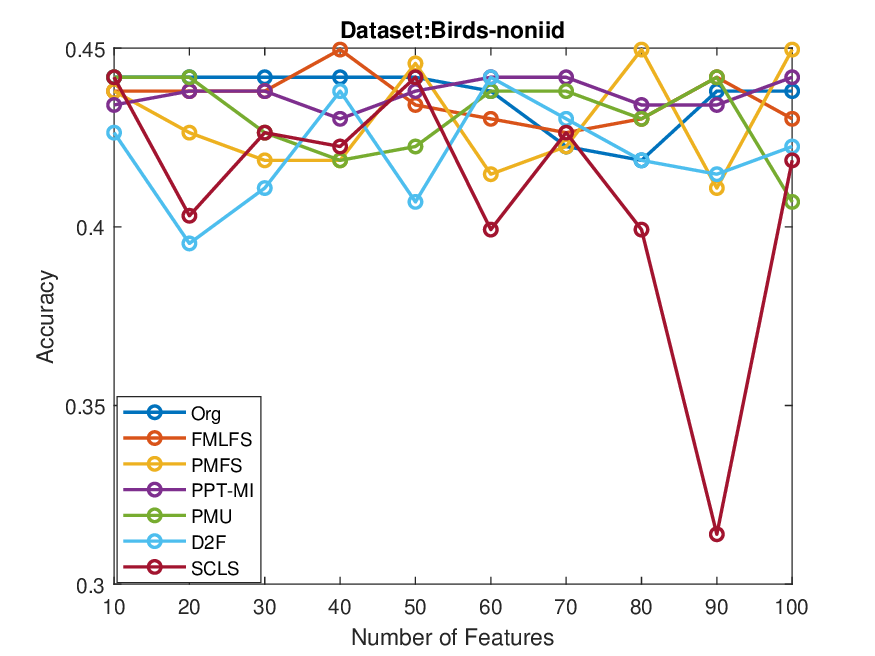}
      \caption{$Accuracy$}
    \end{subfigure}&
    \begin{subfigure}{0.22\textwidth}
      \includegraphics[width=\textwidth]{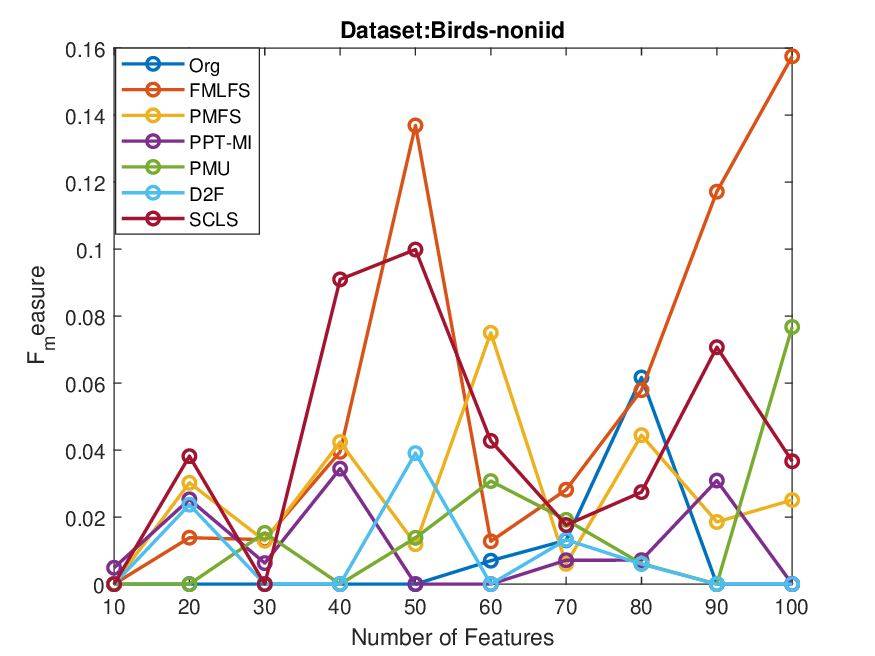}
      \caption{$F-measure$}
    \end{subfigure}\\
    \begin{subfigure}{0.22\textwidth}
      \includegraphics[width=\textwidth]{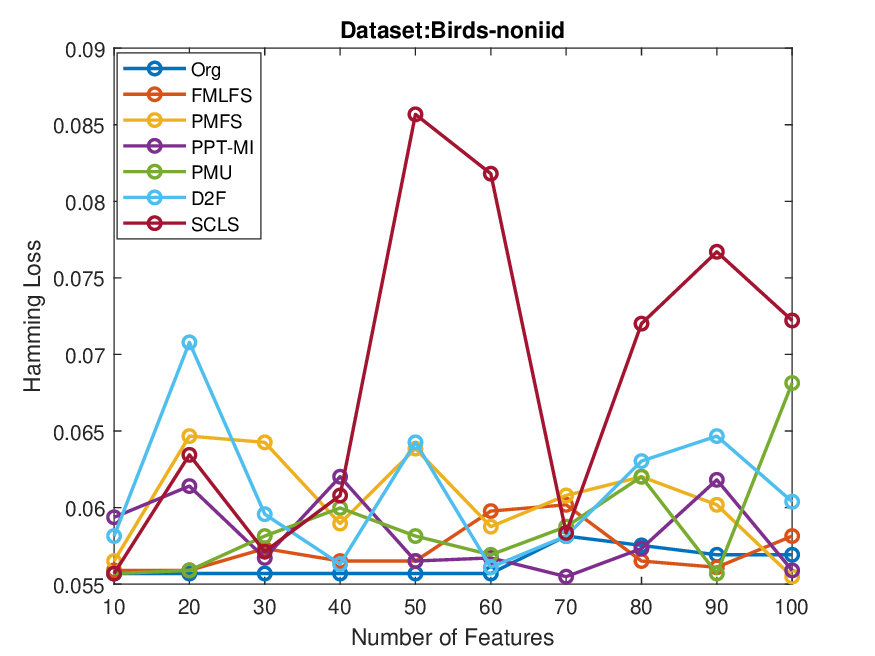}
      \caption{$Hamming Loss$}
    \end{subfigure}&
    \begin{subfigure}{0.22\textwidth}
      \includegraphics[width=\textwidth]{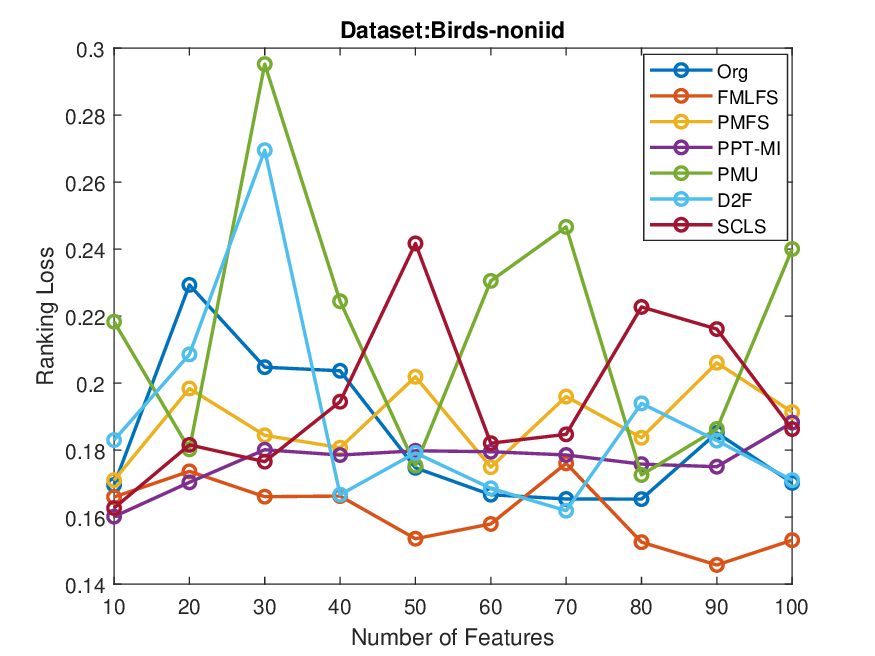}
      \caption{$Ranking Loss$}
    \end{subfigure}\\
    \begin{subfigure}{0.22\textwidth}
      \includegraphics[width=\textwidth]{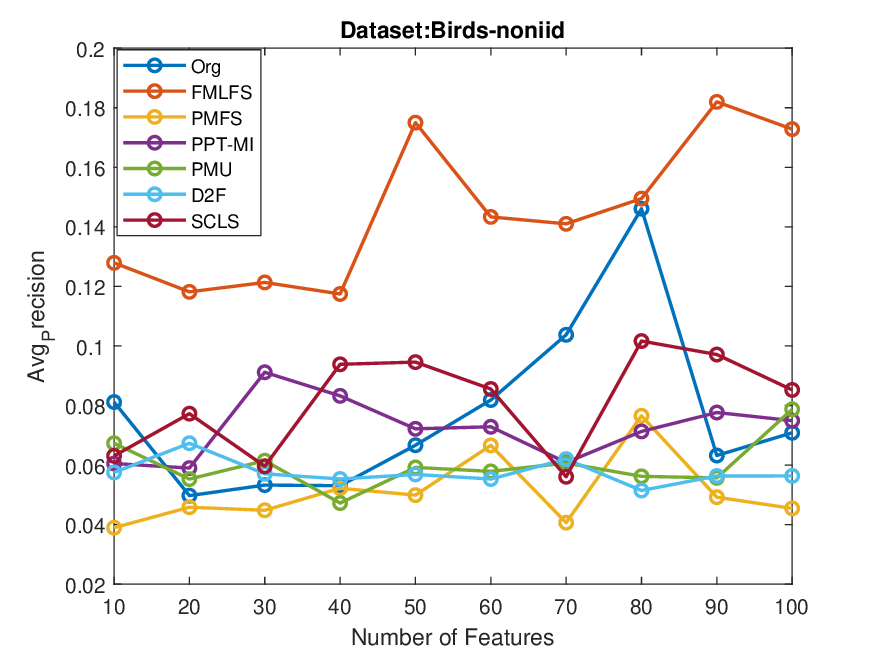}
      \caption{$Avg Precision$}
    \end{subfigure}&
    \begin{subfigure}{0.22\textwidth}
      \includegraphics[width=\textwidth]{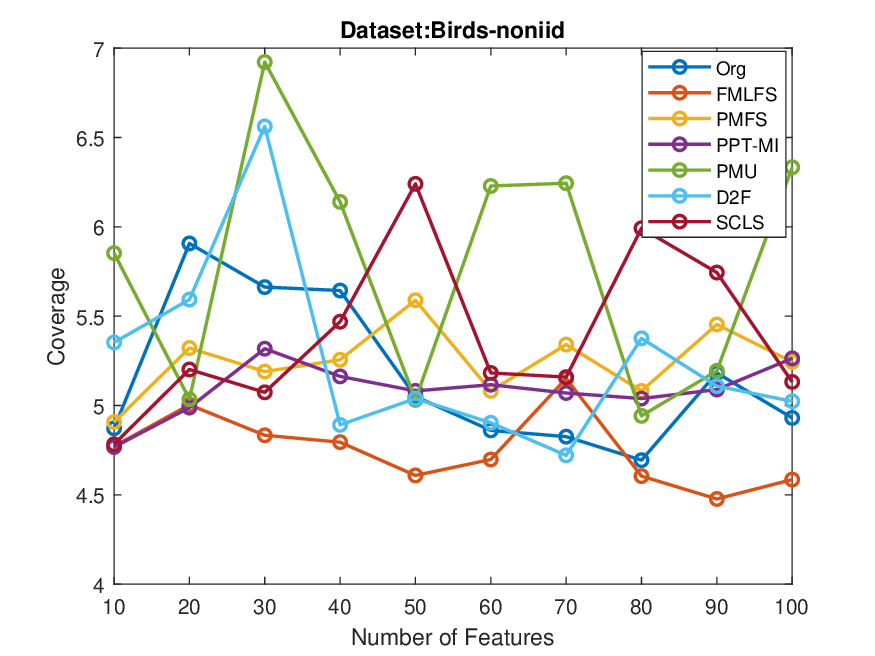}
      \caption{$Coverage$}
    \end{subfigure}
  \end{tabular}
  \vspace{-3mm}
  \caption{Results for Birds non-iid dataset with FL.}\label{fig:animals}
\end{figure}

\begin{figure}[htbp]
  \centering
  \begin{tabular}[c]{cc}
    \begin{subfigure}{0.22\textwidth}
      \includegraphics[width=\textwidth]{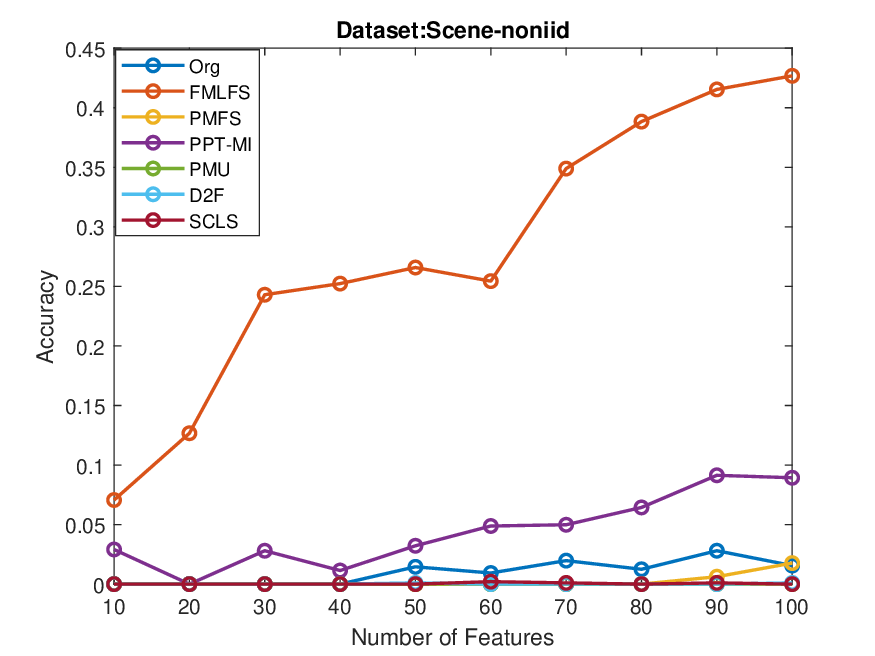}
      \caption{$Accuracy$}
    \end{subfigure}&
    \begin{subfigure}{0.22\textwidth}
      \includegraphics[width=\textwidth]{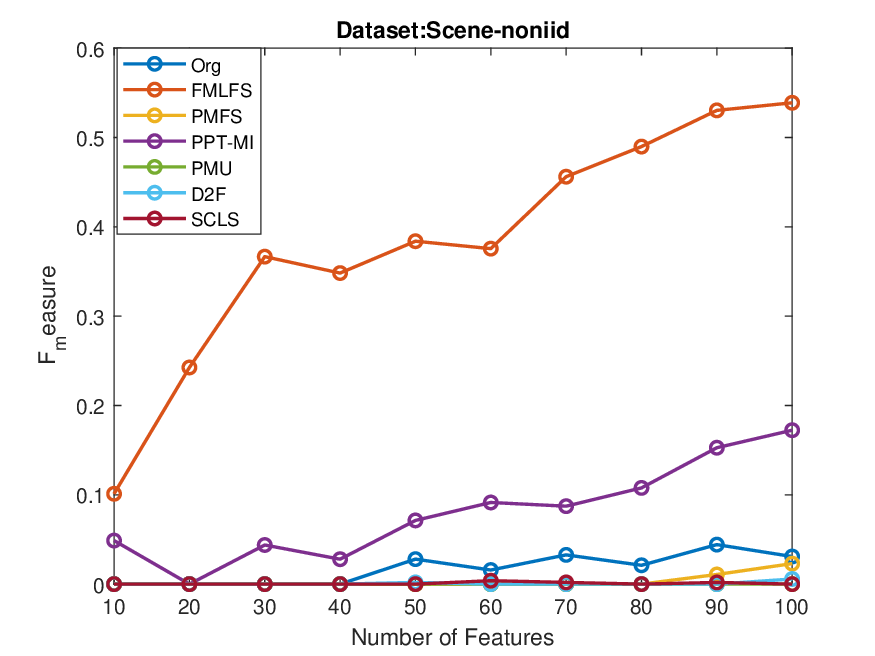}
      \caption{$F-measure$}
    \end{subfigure}\\
    \begin{subfigure}{0.22\textwidth}
      \includegraphics[width=\textwidth]{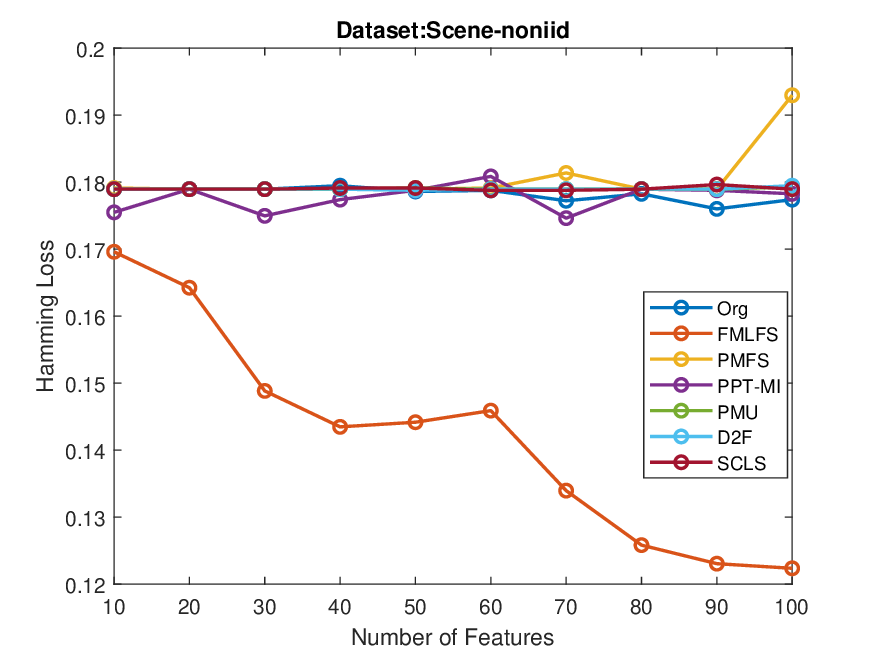}
      \caption{$Hamming Loss$}
    \end{subfigure}&
    \begin{subfigure}{0.22\textwidth}
      \includegraphics[width=\textwidth]{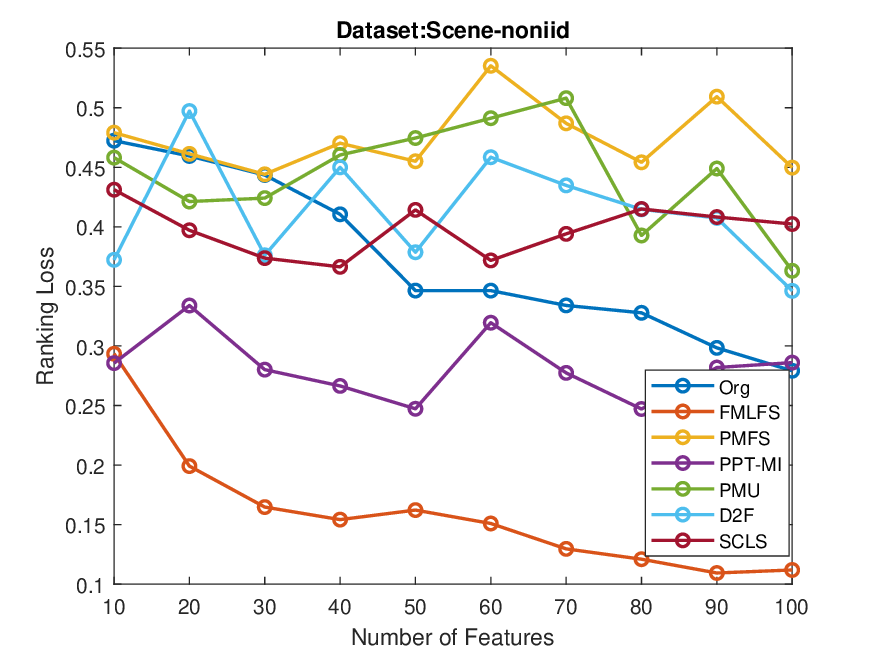}
      \caption{$Ranking Loss$}
    \end{subfigure}\\
    \begin{subfigure}{0.22\textwidth}
      \includegraphics[width=\textwidth]{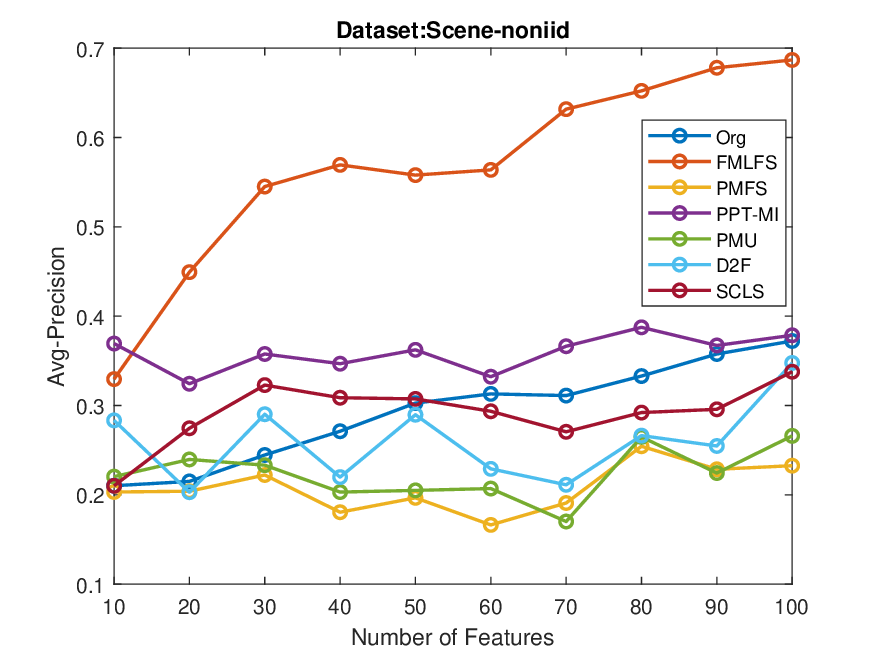}
      \caption{$Avg Precision$}
    \end{subfigure}&
    \begin{subfigure}{0.22\textwidth}
      \includegraphics[width=\textwidth]{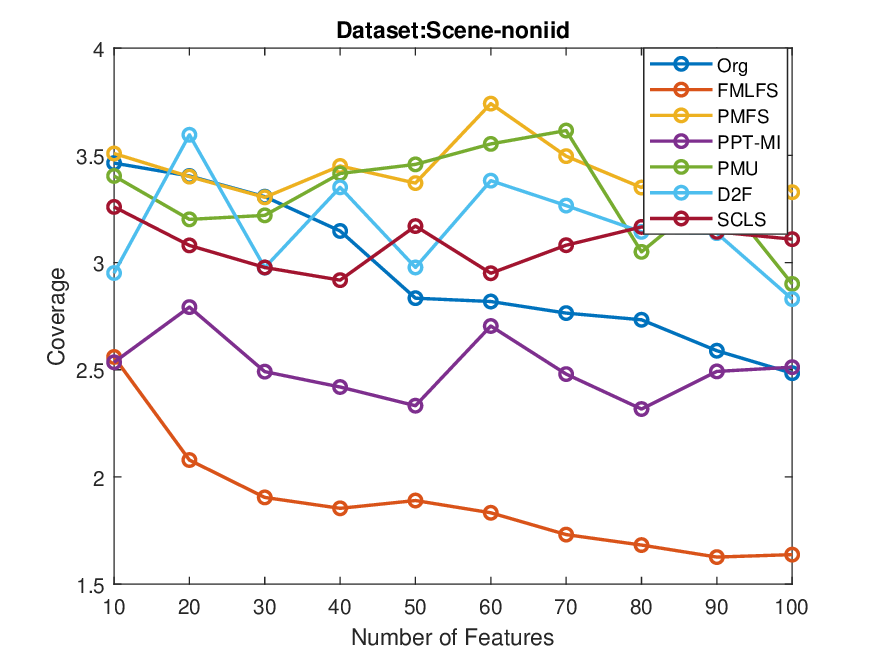}
      \caption{$Coverage$}
    \end{subfigure}
  \end{tabular}
  \vspace{-3mm}
  \caption{Results for Scene non-iid dataset with FL.}\label{fig:animals}
\end{figure}

\vspace{-2mm}
\section{Conclusion and Future Works}
In this paper, we introduce FMLFS, the first federated multi-label feature selection method. Inspired by federated learning, FMLFS comprises two phases. Firstly, within each client, redundancy of features and relevancy between features and labels are computed based on information theory concepts. Subsequently, upon aggregating the received information from clients at the edge server, the multi-label feature selection task is transformed into a bi-objective optimization problem. Utilizing Pareto-based dominance and crowding distance strategies, features are ranked, and the rankings are sent back to the clients. Finally, users can select the desired number of features based on their application requirements. Then, three real-world datasets are utilized to assess both federated learning and centralized learning algorithms, evaluating the performance of the proposed method. The results demonstrate the ability of the proposed method to achieve a good trade-off between performance, time complexity and communication cost. For instance, in the Yeast dataset, the proposed method achieves superior accuracy by selecting just 10 features compared to other methods using 100 features. As we propose a filter-based method in this study, our future work entails integrating federated learning procedures and embedded feature selection methods for distributed multi-label datasets.
\vspace{-2mm}
\section*{Acknowledgement}
This work is funded by research grant provided by the National Science Foundation (NSF) under the grant number 2340075.
\vspace{-2mm}
\bibliographystyle{IEEEtranN}
\bibliography{References}
\end{document}